\documentclass[10pt,twocolumn,letterpaper]{article}

\usepackage[pagenumbers]{cvpr} 

\usepackage{graphicx}
\usepackage{amsmath}
\usepackage{amssymb}
\usepackage{booktabs}
\usepackage{diagbox}
\usepackage{multirow}
\usepackage{makecell}
\usepackage{tabularx}
\usepackage{marvosym}
\usepackage{verbatim}
\usepackage[accsupp]{axessibility} 
\usepackage[dvipsnames]{xcolor}

\definecolor{cvprblue}{rgb}{0.21,0.49,0.74}
\usepackage[pagebackref,breaklinks,colorlinks,citecolor=cvprblue]{hyperref}

\def\paperID{5984} 
\def\confName{CVPR}
\def\confYear{2024}

\title{ZePT: Zero-Shot Pan-Tumor Segmentation via  \\ Query-Disentangling and Self-Prompting}
\author{Yankai Jiang\textsuperscript{1*}, \quad Zhongzhen Huang\textsuperscript{1,2*}, \quad Rongzhao Zhang\textsuperscript{1}, \\  Xiaofan Zhang\textsuperscript{1,2\Letter}, \quad Shaoting Zhang\textsuperscript{1,3\Letter} \\
\textsuperscript{1}Shanghai AI Laboratory \quad
\textsuperscript{2}Shanghai Jiao Tong University \quad \textsuperscript{3}SenseTime Research \\
\tt\small{jiangyankai@pjlab.org.cn, huangzhongzhen@sjtu.edu.cn, zhangrongzhao@pjlab.org.cn,}\\
\tt\small{xiaofan.zhang@sjtu.edu.cn, zhangshaoting@pjlab.org.cn}
}

\begin{document}
\maketitle
\renewcommand{\thefootnote}{\fnsymbol{footnote}}
\footnotetext[1]{Equal contribution. \textsuperscript{\Letter}Corresponding authors. This work was supported by Shanghai Artificial Intelligence Laboratory. Codes are available at https://github.com/Yankai96/ZePT.
}

\begin{abstract}

The long-tailed distribution problem in medical image analysis reflects a high prevalence of common conditions and a low prevalence of rare ones, which poses a significant challenge in developing a unified model capable of identifying rare or novel tumor categories not encountered during training. In this paper, we propose a new \textbf{Ze}ro-shot \textbf{P}an-\textbf{T}umor segmentation framework (ZePT) based on query-disentangling and self-prompting to segment unseen tumor categories beyond the training set. ZePT disentangles the object queries into two subsets and trains them in two stages. Initially, it learns a set of fundamental queries for organ segmentation through an object-aware feature grouping strategy, which gathers organ-level visual features. Subsequently, it refines the other set of advanced queries that focus on the auto-generated visual prompts for unseen tumor segmentation. 
Moreover, we introduce query-knowledge alignment at the feature level to enhance each query's discriminative representation and generalizability. Extensive experiments on various tumor segmentation tasks demonstrate the performance superiority of ZePT, which surpasses the previous counterparts and evidences the promising ability for zero-shot tumor segmentation in real-world settings. 


\end{abstract}

\section{Introduction}
\label{sec:intro}

\begin{figure}
  \centering
  \includegraphics[width=\linewidth]{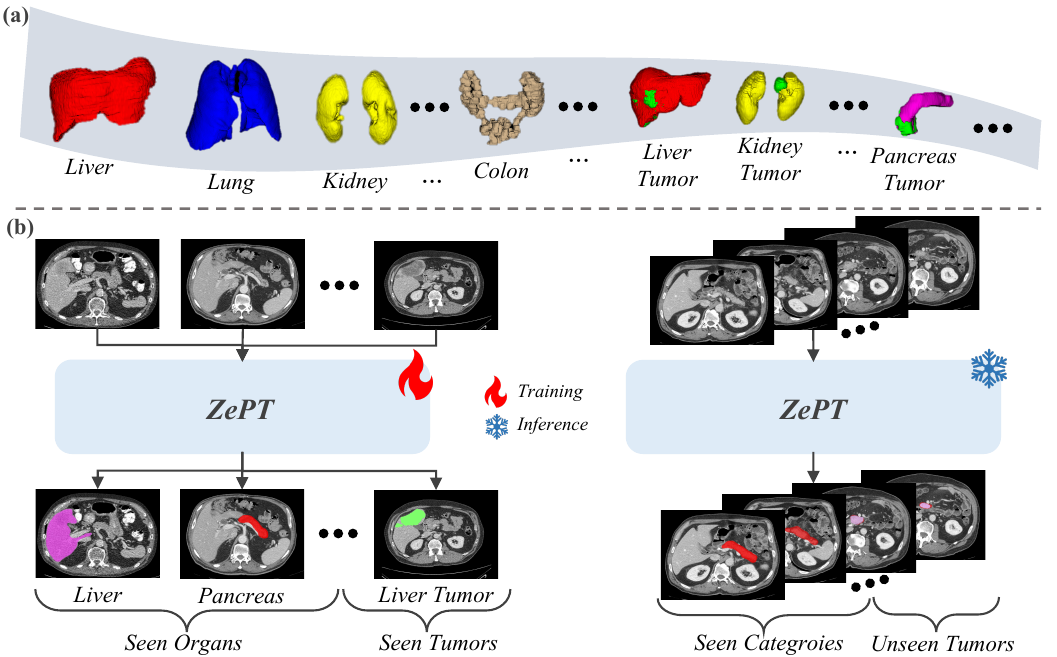}
  \caption{
  (a) The long-tailed distribution issue in medical image analysis. 
  (b) ZePT is trained on datasets containing multiple organs and tumors. During inference, ZePT can segment both seen categories (\ie organs and tumors) and unseen tumors.
  }
  \vspace{-10pt}
  \label{fig:motivation}
\end{figure}

A key challenge in medical image analysis stems from the long-tailed distribution problem, characterized by heavily imbalanced datasets where a few common cases coexist with many rare diseases~\cite{zhang2023challenges} (\cref{fig:motivation} (a)). 
Most existing methods trained on specific-purpose datasets solely focus on a narrow scope of organs or tumors~\cite{seo2019modified,fan2020inf,heller2021state,zhao20213d,qu2023transformer,ding2021rfnet,jiang2021ala}.
Recently, some studies~\cite{liu2023clip,chen2023towards} attempted to design general-purposed methods that can handle various organs and tumors with a unified model. However, these models require large amounts of labeled training data and still have difficulty in identifying rare or new lesion categories that are clinically relevant. Obtaining gold-standard annotation for every tumor category from clinical experts can be highly expensive due to labor-intensive manual efforts, complex annotation processes~\cite{hosny2018artificial}, and may incur privacy concerns. 
In such a scenario, a zero-shot segmentation approach is highly desired, where the model can automatically segment unseen diseases without prior exposure to annotated cases during training. Therefore, we aim to explore the potential of zero-shot segmentation in developing a general-purpose medical image segmentor, as illustrated in~\cref{fig:motivation} (b). 

The zero-shot segmentation~\cite{bucher2019zero} paradigm has been widely studied in the general image processing field~\cite{xian2019semantic,zhang2021prototypical,baek2021exploiting,gu2020context}, which replaces the learnable weights of the classifier with fixed class semantic embeddings~\cite{mikolov2013distributed,joulin2016bag,pennington2014glove,devlin2018bert} to transfer knowledge from seen (base) categories to unseen (novel) ones. Nevertheless, the performance of these methods is bottlenecked due to the absence of necessary knowledge about the novel classes~\cite{wu2023towards, zhu2023survey}. In more recent developments, open-vocabulary semantic segmentation (OVSS) techniques~\cite{ding2022decoupling,liang2023open,ghiasi2022scaling,xu2022simple,qin2023freeseg} utilize vision-language models (VLMs), \eg, CLIP~\cite{radford2021learning}, to significantly enhance the accuracy of zero-shot segmentation. 
Some of them further utilize object queries from MaskFormer~\cite{cheng2021per,cheng2022masked} trained on base categories to produce class-agnostic mask proposals and then classify proposals with VLMs, demonstrating strong and robust zero-shot segmentation capabilities.

Although OVSS methods have achieved success in segmenting novel categories, their performance heavily depends on the quality of the generated proposals. Our pilot experiments in~\cref{tab:MSD} indicate a primary challenge in applying the conventional OVSS strategy to tumor segmentation tasks: the vision clues of the semantics of tumors are usually more subtle and ambiguous than most common-life objects in natural images, which, as a consequence, makes it difficult to generate high-quality proposals for the unseen tumor categories. Therefore, the commonly adopted assumption in most OVSS methods~\cite{ding2022decoupling,liang2023open,ghiasi2022scaling,xu2022simple,qin2023freeseg} that the generated proposals cover almost all the potential object-of-interest no longer holds in the scenario of tumor segmentation, where a considerable number of tumors of unseen categories may be poorly covered.

Driven by the aforementioned limitations, we present a novel framework named ZePT for zero-shot tumor segmentation. 
ZePT adopts a query-disentangling scheme that partitions the object queries into two distinct subsets: fundamental queries and advanced queries. Then we decouple the learning process into two stages, allowing the model to first understand comprehensive organ anatomies and then focus on tumor segmentation, analogous to the learning process of human radiologists.
In Stage-I, we pretrain the fundamental queries via object-aware feature grouping to acquire organ-level semantics for precise organ segmentation. In Stage-II, we train advanced queries, guided by self-generated visual prompts emerging from the fundamental queries, to concentrate on the subtle visual cues associated with tumors. 
Through the query-disentangling and self-prompting, ZePT captures fine-grained visual features associated with pathological changes and generates high-quality proposals that precisely cover unseen tumors. At last, we introduce cross-modal alignment between automatically sourced medical domain knowledge and query embeddings to provide weak supervision and augment the model with additional high-level semantic information, further enhancing the model's generalizability to unseen tumors.



In our experiments, we train ZePT using an assembly of $10$ public benchmarks. We measure the tumor segmentation performance on MSD dataset~\cite{antonelli2022medical} and a curated real-world dataset in a zero-shot manner. ZePT shows robust segmentation performance across four unseen tumor categories, significantly outperforming the previous leading methods by an average improvement of $15.85\%$ in DSC, $17.43\%$ in AUROC, and $23.27\%$ in FPR$_{95}$.
Meanwhile, ZePT also improves the segmentation performance of seen organs and tumors by at least absolute $4.83\%$ in DSC on BTCV~\cite{landman2015miccai}, $4.51\%$ in DSC per case score on LiTS~\cite{bilic2023liver}, $2.21\%$ in DSC on KiTS~\cite{heller2020international}, compared with the strong baseline nnUNet~\cite{isensee2021nnu} and Swin UNETR~\cite{tang2022self}. 

Our main contributions can be summarized as follows:
\begin{itemize}
\item We propose ZePT, a novel two-stage framework with a query-disentangling scheme tailored for zero-shot tumor segmentation. 
\item We formulate tumor segmentation as a unique self-prompting process to localize unseen tumors.
\item ZePT performs feature-level alignment between object queries and medical domain knowledge, further enhancing its generalizability to unseen tumors.
\item ZePT consistently outperforms SOTA counterpart methods on multiple segmentation tasks, showing its effectiveness and robustness. 
\end{itemize}

\section{Related Work}
\label{sec:relatedwork}

\noindent \textbf{Multi-Organ and Tumor Segmentation.}
The advancement of innovative model architectures~\cite{isensee2021nnu,yu2020c2fnas,fang2020multi,zhang2021dodnet} and learning strategies~\cite{shi2021marginal,tang2022self,xie2022unimiss,zhou2023unified,ji2023continual,jiang2023anatomical} has significantly propelled the field of automatic multi-organ segmentation, allowing it to achieve expert-level performance. Despite this significant progress, pan-tumor segmentation persistently presents a challenge. Existing efforts are usually specialized for single tumors~\cite{jiang2018multiple,heller2021state,hatamizadeh2021swin,jiang2021ala,yao2022deepcrc,qu2023transformer}. Some latest attempts are dedicated to training a universal model for segmenting various organs and tumors~\cite{liu2023clip,chen2023towards}. In addition, a growing trend is emerging in efforts~\cite{ma2023segment,wu2023medical,cheng2023sam,chen2023ma,wang2023sam} to transfer the capabilities of SAM~\cite{kirillov2023segment} to segment the abdominal organs and specific tumors. Differently, ZePT takes one step further by investigating a model that 
is capable of segmenting tumors from multiple organs in a zero-shot manner.

\noindent \textbf{Open-Vocabulary Semantic Segmentation.}
The emerging concept of OVSS defines a generalized zero-shot semantic segmentation paradigm that allows a model to be trained on conventional vision datasets with close-set labels while possessing the ability to segment an image into arbitrary semantic regions according to text descriptions~\cite{ding2022decoupling,liang2023open,ghiasi2022scaling,xu2022simple,qin2023freeseg}. Nevertheless, as pointed out in~\cite{liang2023open}, the mask proposals in OVSS methods are not truly ``class-agnostic''. They tend to overfit to seen categories and fail to cover previously unseen obscure objects. This issue hinders the transfer of the power of OVSS to zero-shot tumor segmentation on medical images. Differently, ZePT disentangles the object queries into two sets and adopts a self-prompting strategy to guide the model to explicitly learn semantics related to unseen (novel) tumor categories.

\begin{figure*}[htbp]
  \centering
  \includegraphics[width=\linewidth]{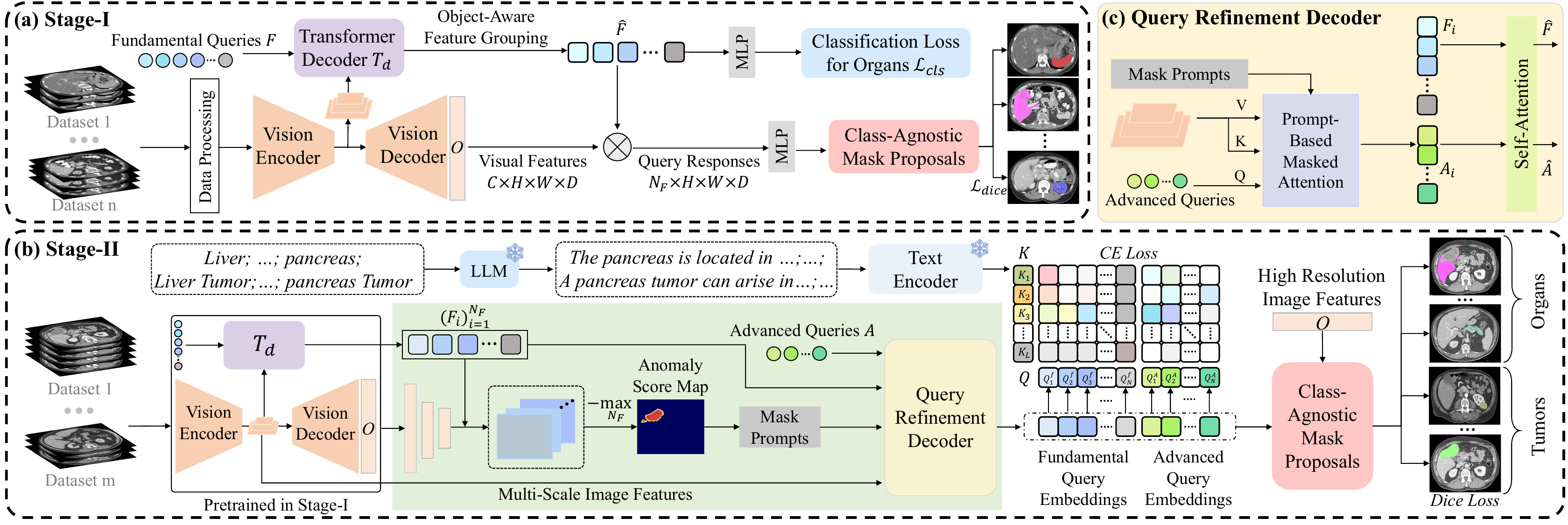}
  \caption{Overall pipeline. \textbf{Stage-I}: Based on MaskFormer~\cite{cheng2021per,cheng2022masked}, we propose an object-aware feature grouping strategy to train a set of fundamental queries for multi-organ segmentation. \textbf{Stage-II}: A set of advanced queries for tumor segmentation attend to visual prompts derived from the affinity between fundamental query embeddings and visual features which indicates the presence of unseen abnormalities. Finally, we incorporate medical domain knowledge to better align text embeddings with query embeddings for cross-modal reasoning.}
  \label{fig:framework}
\end{figure*}

\noindent \textbf{Unseen Lesion Detection and Segmentation.}
Some research efforts~\cite{tian2021constrained,pinaya2022unsupervised,zimmerer2022mood,roy2022does,yuan2023devil} also explored innovative approaches to detect or segment unseen lesions/tumors, which formulated the task as out-of-distribution (OOD) detection. 
Although unseen tumors can be regarded as a kind of OOD sample, the zero-shot segmentation task in this paper is largely different from OOD detection from the perspective that our main objective is to segment and classify multi-class tumors in a zero-shot manner with text descriptions, while OOD detection only recognizes unseen tumors as one single outlier class. 

\section{Method}

As illustrated of~\cref{fig:framework}, ZePT differs significantly from existing OVSS approaches~\cite{ding2022decoupling,liang2023open,ghiasi2022scaling,xu2022simple,qin2023freeseg} in that it disentangles the learning process of seen organs and unseen tumors into two stages and partitions object queries into two distinct subsets: fundamental queries and advanced queries. Specifically, Stage-I aims at pretraining fundamental queries on multiple datasets containing only organ labels to attain high-quality organ segmentation capability. In Stage-II, we formulate tumor segmentation as a self-prompting process, where advanced queries attend to the visual prompts derived from the affinity between embeddings of fundamental queries and visual features for capturing critical fine-grained context information and pathological changes related to tumors. We elaborate on the details of our designs in the following.


\subsection{Stage-I: Fundamental Queries for Organs} \label{BQGMP}
We build our model upon a MaskFormer~\cite{cheng2021per,cheng2022masked}. As shown in~\cref{fig:framework} (a), the segmentation backbone consists of three components. A vision encoder $V_e$ that extracts multi-scale visual features $V=\{V_i\}^4_{i=1}$, $ V_i \in \mathbb{R}^{H_i \times W_i \times D_i \times C_i}$ from 3D volumes. Here, $H_i$, $W_i$, $D_i$ and $C_i$ denote the height, width, depth and channel dimension of $V_i$, respectively. A transformer decoder $T_d$ that updates a set of $N_F$ learnable fundamental queries $F\in\mathbb{R}^{N_F \times C}$ by interacting them with multi-scale visual features. A vision decoder $V_d$ that gradually upsamples visual features to high-resolution image embeddings $O \in \mathbb{R}^{ H \times W \times D \times C}$. 

To capture organ-level information and achieve object-aware cross-modal reasoning, we propose an object-aware feature grouping strategy in $T_d$ to guide each learnable fundamental query to represent and specify a corresponding organ category. This is achieved by grouping visual features into the query embeddings for context-aware reasoning. 

Specifically, $T_d$ consists of a series of transformer blocks enabling the queries to interact with multi-scale features. 
In the $i$-th transformer block, the fundamental queries $F\in\mathbb{R}^{N_F \times C}$ first exploit global information from 3D image feature maps $ V_i \in \mathbb{R}^{H_i \times W_i \times D_i \times C_i}$ via a classical cross-attention as follows:
\begin{gather}
\mathcal{Q} = W^q  \delta(F), \mathcal{K}=W^k  V_i, \mathcal{V}= W^v  V_i \\
\hat{F}_i = \operatorname{MLP}(\operatorname{LayerNorm}(\operatorname{Softmax}\left(\frac{\mathcal{Q} \mathcal{K}^T}{\sqrt{d}}\right) \mathcal{V}),
\end{gather}
where $ W^q, W^k, W^v \in \mathbb{R}^{C_i \times C_i}$ are learnable projection matrices. $\delta$ is a linear projection.

Subsequently, we explicitly assign the relevant local context information from visual features into the fundamental queries based on affinity in the embedding space to ensure that different queries focus on different visual regions without overlaps.
We first calculate an assignment similarity matrix $S_i\in \mathbb{R}^{N_F\times H_iW_iD_i}$ between the fundamental queries $\hat{F}_{i}$ and the image features $V_i$ via a $\operatorname{Gumbel-Softmax}$~\cite{jang2016categorical,maddison2016concrete} operation:
\begin{gather}
S_i^{\text{gumbel}}=\operatorname{Softmax}\left(\left(\hat{F}_{i}V_i^T+G\right) / \tau\right).
\end{gather}
Here $G \in \mathbb{R}^{N_F\times H_iW_iD_i}$ are i.i.d random samples drawn from the $Gumbel(0,1)$ distribution and $\tau$ is a learnable coefficient to assist in finding a suitable assignment boundary. 

We then group the visual features in $V_i$ and corresponds the groups to the queries $\hat{F}_{i}$ by taking the one-hot operation of the $\operatorname{argmax}$ over $S_i^{\text{gumbel}}$:
\begin{gather}
S_i^{\text{onehot}}=\operatorname{onehot}\left(\operatorname{argmax}_{N_F}\left(S_{\text{gumbel}}\right)\right) .
\end{gather}
Since the straightforward hard assignment (\ie, one-hot) via $\operatorname{argmax}$ is not differentiable, we adopt the straight through trick in~\cite{van2017neural,xu2022groupvit} to compute the assignment similarities $\hat{S}_{i}$ of one-hot value as follows:
\begin{equation}
 \hat{S}_{i}=\left(S_i^{\text{onehot}}\right)^{\top}+S_i^{\text{gumbel}}-\operatorname{sg}\left(S_i^{\text{gumbel}}\right),
\end{equation}
where $\operatorname{sg}$ is the stop gradient operator. 

With the above operations, the whole transfomrer block is differentiable and end-to-end trainable. $\hat{S}_{i}$ indicates the assignment of object-level visual features to each query. Finally, query embeddings $\hat{F}_{i}$ will be updated via being assigned with the most corresponding features in $V_i$ according to $\hat{S}_{i}$, which can be denoted as follows:
\begin{equation}
\hat{F}_{i+1}=\operatorname{MLP}\left(S_i V_i\right)+\hat{F}_i .
\end{equation}


The binary mask proposals $BM \in [0, 1]^{N_F \times H\times W \times D}$ for fundamental queries are obtained by a multiplication operation between the query embedding and high-resolution image features $O$ followed by a Sigmoid. We adopt Dice Loss to supervise mask proposals with organ labels. We also process the query embeddings through a $\operatorname{MLP}$ layer to get class embeddings, which are then supervised using the category information of organs through a Cross-Entropy loss. As later shown in~\cref{fig:ATTvis} (a), our learnable fundamental queries focus on distinct foreground organ regions and explicitly encourage the strict boundaries between different categories, preventing mixed representations where the target region and the disturbing regions are grouped together. Such discriminative representation also enhances the localization of unseen tumors, as discussed in~\cref{PUQ}. 

\subsection{Stage-II: Advanced Queries for Tumors} \label{PUQ}
In Stage-II, we aim to refine a set of $N_A$ advanced queries $A \in \mathbb{R}^{N_A \times C}$ for tumor segmentation and ensure their generalization to unseen tumors. The core insight is to endow the advanced queries with the ability to capture anomaly information of tumors by utilizing fundamental queries. 
We propose to reformulate tumor segmentation as a self-prompting process where the advanced queries can be aware of abnormal information related to pathological changes in the feature context via visual prompts. We retain the training datasets in Stage-I to avoid the forgetting problem of the fundamental queries and add several datasets containing tumor labels for Stage-II. It is worth noting that there are novel tumor categories in the testing phase, rendering our method truly ``zero-shot''.


\textbf{Self-Generated Visual Prompts.}
As shown in~\cref{fig:framework} (b), we utilize the pretrained $V_e$, $V_d$, $T_d$, and fundamental queries $F$ from Stage-I. 
The volumes are fed into the pretrained network to obtain the multi-scale visual features $V=\{V_i\}^4_{i=1}$, high-resolution image embeddings $O \in \mathbb{R}^{ H \times W \times D \times C}$ and refined fundamental query embeddings $\hat{F} \in \mathbb{R}^{ N_F\times C}$. 
We compute multi-scale query response maps $R_i \in \mathbb{R}^{N_F \times H_i \times W_i \times D_i}$ representing the affinity between visual features and different fundamental queries at each resolution stage. 
Then we adopt the negative of maximal operation~\cite{yuan2023devil} along channel dimension on $R_i$ to generate multi-scale anomaly score $M_i \in \mathbf{R}^{H_i \times W_i \times D_i}$ maps:
\begin{gather}
    R_i = \hat{F}_iV_i^T \\
    M_i = \mathop{-\operatorname{max}}\limits_{c \in {1, \dots, N_F}} R_i^c, i \in [1,4].
\end{gather}

These anomaly score maps $M_i$ can be further normalized into mask prompts $\hat{M}_i \in [0, 1]^{H_i\times W_i \times D_i}$ by min-max normalization and a threshold of $0.5$, where $\hat{M}_i^{(h,w,d)} = 1$ and $\hat{M}_i^{(h,w,d)} = 0$ represent that the voxel located at position $(h,w,d)$ in the input 3D volume belongs to an anomalous unseen category and an in-distribution seen organ class, respectively.
We use these mask prompts, adaptively derived from the embedding space, to assist the advanced queries to attend to the anomalous context features and learn representations that effectively localize unseen tumors.

\textbf{Query Refinement Decoder.}
The Query Refinement Decoder (QRD) takes mask prompts, multi-scale visual features, a set of zero-initialized advanced queries, and embeddings of fundamental queries as inputs. As shown in~\cref{fig:framework} (c), the $N_A$ advanced queries are designed to localize and identify unseen tumors on organs corresponding to the $N_F$ fundamental queries. For the $i$-th block in QRD, the advanced queries $A_i$ are first updated via interactions with multi-scale features $V_i$ and mask prompts $\hat{M}_i$ via:
\begin{gather}
  \hat{A}_{i} = {A_i} + \operatorname{Softmax}(\mathcal{M}_i + \mathcal{Q}_{A_i}\mathcal{K}_{V_i}^T){\mathcal{V}_{V_i}}^T,
  \label{eq:2}
\end{gather}
where $\mathcal{Q}_{A_i} = f_Q(A_i) \in \mathbb{R}^{N_A \times C_i}$ denotes embeddings of advanced queries under transformation $f_Q(\cdot)$. $\mathcal{K}_{V_i}$, $\mathcal{V}_{V_i} \in \mathbb{R}^{C_i \times H_iW_iD_i}$ denote 3D image features under transformation $f_{\mathcal{K}}(\cdot)$ and $f_{\mathcal{V}}(\cdot)$, respectively. The visual prompt attention mask $\mathcal{M}_i$ at feature location $(h,w,d)$ is defined as:
\begin{equation}
  \mathcal{M}_i^{(h,w,d)} = 
  \begin{cases}
       0 & \mbox{if} \: \hat{M}_i^{(h,w,d)} = 1\\
       -\infty & \mbox{otherwise}
  \end{cases}
  .
 \label{eq:3}
\end{equation}
We refer to this mechanism as \textbf{prompt-based masked attention} because it aggregates the visual information highlighted by given mask prompts, allowing advanced queries to concentrate on abnormal regions with pathological changes across various organs, thereby facilitating the detection of previously unseen tumors. Through a series of blocks, we can obtain the refined advanced queries $\hat{A}_i$. Then, QRD concatenates fundamental queries $\hat{F}_i$ with advanced queries $\hat{A}_i$ and performs self-attention among them to 
encode the relationship between fundamental and advanced queries, facilitating the adjustment of their representations to encourage a clear semantic distinction between organs and tumors. Finally, we use these updated queries to generate corresponding mask proposals as described in~\ref{BQGMP}. Organ labels supervise the mask proposals derived from the fundamental queries, whereas tumor labels supervise those from advanced queries. 

\textbf{Query-Knowledge Alignment.}
To make the model better retain its generalization ability for recognizing unseen tumors, we introduce Query-Knowledge Alignment for the weakly-supervised cross-modal alignment between visual features of queries and high-level semantics of textual knowledge. Instead of a simple description (\eg ``a photo of {}'') in previous methods~\cite{ding2022decoupling,liang2023open,xu2022simple,liu2023clip}, we utilize detailed knowledge of each class name by prompting GPT4~\cite{openai2023gpt4} with an instruction: ``Please describe \{CLS\}. The answer should encompass attributes related to its location, shape, size, and anatomical structure.''. Each piece of generated knowledge is checked and modified by professional doctors to ensure correctness. We adopt ClinicalBERT~\cite{alsentzer2019publicly} as a pretrained text encoder to get the knowledge embeddings $\mathbf{K}$. The predicted probability distribution over the training classes for the $i$-th query is calculated as:
\begin{equation}
  \mathcal{P}_i = \frac{\mbox{exp}(\frac{1}{\tau}\zeta(\mathbf{K}_i, \mathbf{Q}_i))}{\sum_{j=0}^{|L|}\mbox{exp}(\frac{1}{\tau}\zeta(\mathbf{K}_j, \mathbf{Q}_i))},
  \label{eq:5}
\end{equation}
where $\zeta$ is the cosine similarity between two embeddings, and $\tau$ is the temperature. $\mathbf{Q}$ is derived from query embeddings via a linear projection layer. During training, the similarities between matched query embedding and text embedding should be maximized. A cross-entropy loss is applied on $\mathcal{P}$ for supervision.

In summary, we combine the Dice loss on mask proposals and the cross-entropy loss for query-knowledge alignment to supervise the learning in Stage-II.

\begin{table*}[htbp]
  \centering
  \resizebox{\linewidth}{!}
  {
  \begin{tabular}{@{}l|ccc|ccc|ccc|ccc|ccc@{}}
    \toprule
    \multicolumn{1}{l}{\multirow{3}{*}{Method}} & \multicolumn{12}{|c|}{MSD Dataset} & \multicolumn{3}{c}{\multirow{2}{*}{\makecell[c]{Real-World Colon\\Tumor Segmentation}}}\\
    \cline{2-13}
     & \multicolumn{3}{c|}{Pancreas Tumor} & \multicolumn{3}{c|}{Lung Tumor} & \multicolumn{3}{c|}{Hepatic Vessel Tumor} & \multicolumn{3}{c|}{Colon Tumor} & \\
    \cline{2-4}\cline{5-7}\cline{8-10} \cline{11-13} \cline{14-16}     
     & AUROC$\uparrow$ & FPR$_{95}$$\downarrow$ & DSC$\uparrow$ & AUROC$\uparrow$ & FPR$_{95}$$\downarrow$ & DSC$\uparrow$
    & AUROC$\uparrow$ & FPR$_{95}$$\downarrow$ & DSC$\uparrow$ & AUROC$\uparrow$ & FPR$_{95}$$\downarrow$ & DSC$\uparrow$
    & AUROC$\uparrow$ & FPR$_{95}$$\downarrow$ & DSC$\uparrow$\\
    \hline
     ZegFormer~\cite{ding2022decoupling}     & 66.45 & 69.33 & 14.92        & 41.31 & 81.78 & 9.94          & 75.39 & 55.94 & 30.81                & 50.13 & 78.81 & 11.34         & 55.64 & 74.29 & 12.03 \\
     zsseg~\cite{xu2022simple}              & 53.23 & 79.27 & 11.40              & 34.96 & 87.54 & 7.98    & 71.43 & 60.35 & 28.57             & 47.79 & 82.68 & 9.68        & 50.30 & 79.52 & 10.18  \\
     OpenSeg~\cite{ghiasi2022scaling}       & 44.56 & 85.19 & 10.05         & 23.49 & 91.75 & 6.12         & 59.23 & 70.52 & 23.38             & 41.76 & 89.44 & 7.13         & 41.92 & 87.51 & 7.59  \\
     OVSeg~\cite{liang2023open}            & 70.22 & 59.73 & 19.36         & 52.93 & 68.65 & 14.11         & 85.77 & 40.28 & 35.66             & 59.94 & 65.25 & 15.76             & 69.95 & 64.84 & 16.05 \\
     FreeSeg~\cite{qin2023freeseg}      & 69.98 & 60.75 & 18.19          & 49.92 & 70.39 & 13.26            & 85.62 & 41.77 & 35.08             & 56.45 & 68.49 & 14.71             & 67.01 & 66.07 & 15.30  \\
    \hline
    SynthCP~\cite{xia2020synthesize}     & 51.24 & 81.69 & 11.33        & 25.85 & 90.28 & 6.43             & 70.12 & 63.55 & 28.01         & 43.84 & 87.71 & 8.74              & 48.37 & 84.95 & 8.72 \\
    SML~\cite{jung2021standardized}     & 37.95 & 89.93 & 9.72            & 20.18 & 93.65 & 6.02          & 57.44 & 70.96 & 22.97         & 22.41 & 92.07 & 6.65              & 39.88 & 88.41 & 7.21   \\
    MaxQuery~\cite{yuan2023devil}      & 68.99 & 59.93 & 18.15          & 48.24 & 70.47 & 11.29             & 83.66 & 42.45 & 34.30         & 50.47 & 69.88 & 13.43              & 64.53 & 67.65 & 15.24 \\
    \hline
    ZePT    & \textbf{86.81} & \textbf{35.18} & \textbf{37.67} 
    & \textbf{77.84} & \textbf{44.30} & \textbf{27.22}
    & \textbf{91.57}   & \textbf{20.64}  & \textbf{52.94} 
    & \textbf{82.36} & \textbf{40.73} & \textbf{30.45}  
    & \textbf{84.35}  &\textbf{38.29}    & \textbf{36.23} \\
    \bottomrule
  \end{tabular}
  }
  \caption{Detection and segmentation performance of unseen tumors on MSD~\cite{antonelli2022medical} and real-world colon tumor dataset. ZePT achieves state-of-the-art unseen tumor detection and segmentation performance. More results can be found in the supplemental material.
  }
  \label{tab:MSD}
\end{table*}

\section{Experiments}
\label{results}

\noindent \textbf{Dataset Construction.}
(1) \textbf{Training:} In {Stage-I}, we assemble the training sets of $8$ public datasets, including Pancreas-CT~\cite{roth2015deeporgan}, AbdomenCT-1K~\cite{ma2021abdomenct}, CT-ORG~\cite{rister2020ct}, CHAOS~\cite{kavur2021chaos}, 
AMOS22~\cite{ji2022amos}, BTCV~\cite{landman2015miccai}, WORD~\cite{luo2022word} and TotalSegmentator~\cite{wasserthal2022totalsegmentator}.
These datasets exclusively contained organ labels. In {Stage-II}, we add CT images from the training sets of LiTS~\cite{bilic2023liver} and KiTS~\cite{heller2020international}. The overall seen categories used for training consist of $25$ organ classes and $2$ tumor classes. (2) \textbf{Inference:} 
We employ the MSD dataset~\cite{antonelli2022medical} that encompasses a range of segmentation tasks for five tumor types in CTs. Among these, pancreas tumors, lung tumors, colon tumors, and hepatic vessel tumors belong to unseen categories. 
A real-world, private dataset containing $388$ 3D CT volumes of four distinct colon tumor subtypes is also utilized for testing.
We follow the data pre-processing in~\cite{liu2023clip} to reduce the domain gap among various datasets. Due to page limits, details of all datasets and pre-processing are described in the supplemental material. 

\noindent \textbf{Evaluation Metrics.} Dice Similarity Coefficient (DSC)
is utilized for evaluating organ/tumor segmentation. We also report the area under the receptive-operative curve (AUROC) 
and the false positive rate at a true positive rate of $95\%$ (FPR$_{95}$), which are commonly used in OOD detection methods~\cite{xia2020synthesize,jung2021standardized,yuan2023devil}.
For all the metrics above, $95\%$ CIs were calculated and the $p$-value cutoff of less than $0.05$ was used for defining statistical significance.

\noindent \textbf{Implementation Details.}
(1) \textbf{Stage-I:} We use the current benchmark model in medical image segmentation, Swin UNETR~\cite{hatamizadeh2021swin}, as the backbone, which consists of a vision encoder and a vision decoder with skip connections. We adopt four transformer decoder blocks in $T_d$, and each takes image features with output stride $32$, $16$, $8$, and $4$, respectively. We employ AdamW optimizer~\cite{loshchilov2017decoupled} with a warm-up cosine scheduler of $50$ epochs. The batch size is set to $2$ per GPU with a patch size of $96 \times 96 \times 96$. The training process uses an initial learning rate of $1e^{-4}$, momentum of $0.9$ and decay of $1e^{-5}$ on multi-GPU ($8$) with DDP for 1000 epochs. Extensive data augmentation is utilized on-the-fly to improve the generalization, including random rotation and scaling, elastic deformation, additive brightness, and gamma scaling.
The number of the fundamental queries $N_F$ is $25$ for $25$ organ classes. 
The loss is the sum of Cross-Entropy loss and Dice loss.
(2) \textbf{Stage-II:} We adopt the pretrained model in Stage-I. We set the initial learning rate as $4e^{-4}$. QRD has four blocks and each attention layer in the QRD block has eight heads. The number of the advanced queries $N_A$ is $20$ for tumors/diseases categories. 
Other settings are kept the same as in Step-I. 
We implement ZePT model in PyTorch~\cite{paszke2019pytorch}. All experiments are conducted on $8$ NVIDIA A$100$ GPUs.

\noindent \textbf{Baselines.} 
In this paper, the zero-shot tumor segmentation setting requires that models directly segment unseen tumor types during inference without any fine-tuning or retraining. This is notably challenging as the model has to handle both unseen classes and domain gaps between different datasets. For unseen tumor segmentation, we compare ZePT with a series of representative OVSS methods, including ZegFormer~\cite{ding2022decoupling}, zsseg~\cite{xu2022simple}, OpenSeg~\cite{ghiasi2022scaling}, OVSeg~\cite{liang2023open}, and Freeseg~\cite{qin2023freeseg}. We also compare ZePT with OOD detection methods~\cite{xia2020synthesize,jung2021standardized,yuan2023devil}, which treat unseen tumors as a single outlier class. 
We adopt the masked back-propagation in~\cite{liu2023clip} to enable the training of these methods on partially labeled datasets. 
All baselines are trained with all datasets used in Stage-I and Stage-II. For seen organ/tumor segmentation, we compare ZePT with SOTA benchmark models, including nnUNet~\cite{isensee2021nnu}, Swin UNETR~\cite{tang2022self} and Universal model~\cite{liu2023clip}. 

\subsection{Main Results}
\noindent \textbf{Unseen Tumor Segmentation on MSD Dataset.}
~\cref{tab:MSD} shows the segmentation performance of four unseen tumor categoreis from MSD~\cite{antonelli2022medical}. All available volumes in these four tumor segmentation tasks are directly used for testing. 
Compared with SOTA OVSS methods, 
ZePT demonstrates a notable performance enhancement in the context of average unseen tumor localization performance across four tasks, achieving at least a $17.43\%$ improvement in AUROC and a $23.27\%$ increase in FPR$_{95}$.
Regarding the average performance in unseen tumor segmentation across these tasks, ZePT continues to maintain a substantial lead, as evidenced by a notable $15.85\%$ improvement in the DSC.
These results indicate that the proposed query-disentangling and self-prompting can effectively help the model capture visual cues related to tumors, thus boosting the ability to recognize unseen ones. Moreover, OVSS methods require an additional frozen CLIP vision encoder to classify each mask proposal, leading to slower inference speeds. In contrast, ZePT removes this process and adopts query-knowledge alignment at the feature level, which maintains a reasonable computation cost.  
As shown in~\cref{tab:Efficiency}, ZePT has fewer network parameters and approximately $23\%$ of the FLOPs compared to previous OVSS methods. 

We also compare ZePT with SOTA OOD detection methods.
ZePT's average performance surpasses the previously best-performing MaxQuery~\cite{yuan2023devil} across four tasks by $21.81\%$ in AUROC, $25.47\%$ in FPR$_{95}$, and $17.78\%$ in DSC. ZePT aligns visual features with linguistic semantics for cross-modal interaction instead of solely exploiting information from visual modality like most OOD region segmentation methods. The results suggest that leveraging features from images together with medical domain knowledge benefits the semantic understanding of unseen tumors.


\noindent \textbf{Real-World Colon Tumor Segmentation.}
We further conduct a zero-shot evaluation on a real-world colon tumor dataset. The average results of four colon tumor subtypes are also summarized in~\cref{tab:MSD}. ZePT outperforms the baselines by at least absolute $14.40\%$ in AUROC, $26.55\%$ in FPR$_{95}$, and $20.18\%$ in DSC, demonstrating much better generalizability and robustness. ZePT reaches a much lower FPR$_{95}$ compared with previous methods, which is crucial for safety-critical medical scenarios.
The results indicate ZePT has a strong potential for utility in clinical practice.

\begin{figure*}[htbp]
  \centering
  \includegraphics[width=\linewidth]{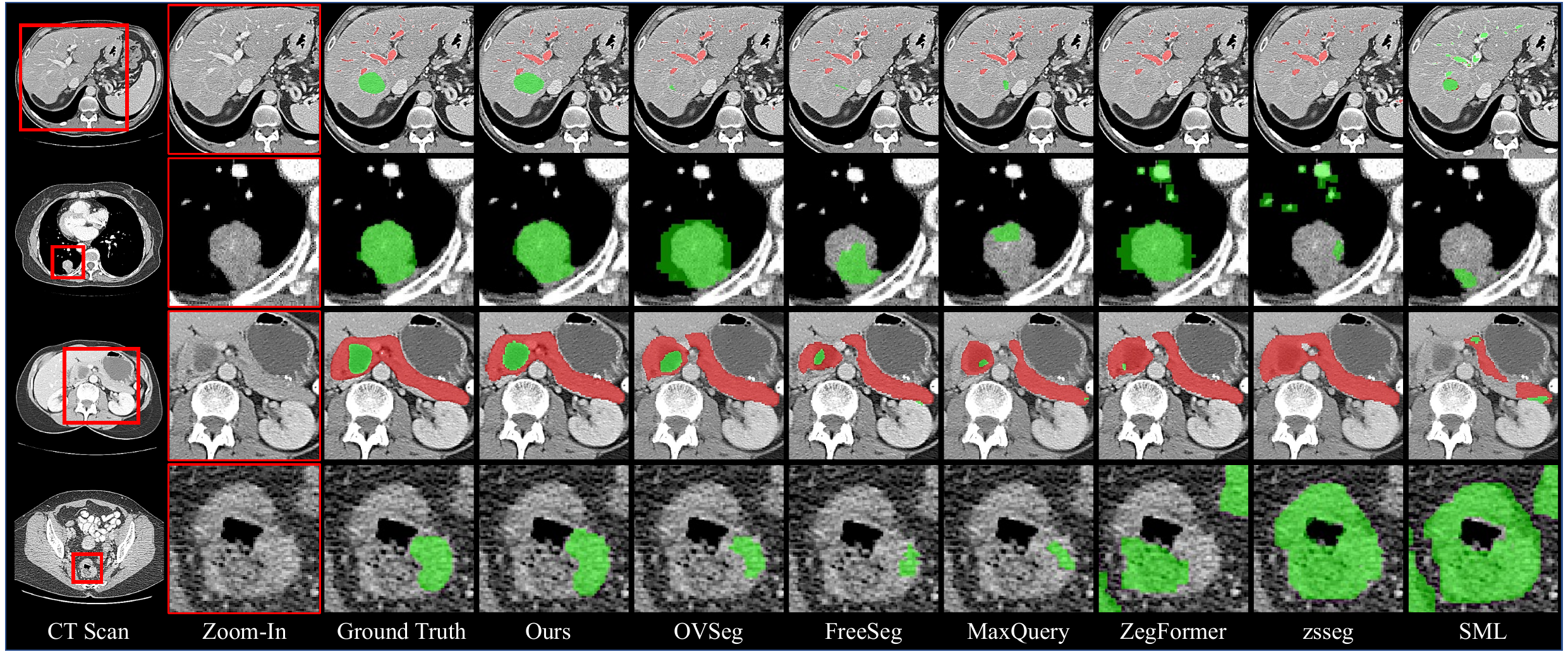}
  \caption{Qualitative visualizations on MSD~\cite{antonelli2022medical} dataset. We compare ZePT with other advanced OVSS methods and OOD detection methods in a zero-shot manner. The segmentation results presented from rows one to four correspond, in order, to hepatic vessel tumors, lung tumors, pancreatic tumors, and colorectal tumors. We present the visualizations on other datasets in the supplemental material.}
  \label{fig:QR}
\vspace{-10pt}
\end{figure*}

\begin{table}
  \Huge
  \centering
  \resizebox{\linewidth}{!}
  {
  \begin{tabular}{@{}c|c|c|c|c@{}}
    \hline
    \diagbox{Efficiency}{Method}  & ZePT & ZegFormer~\cite{ding2022decoupling}  & OVSeg~\cite{liang2023open} & FreeSeg~\cite{qin2023freeseg}\\
    \hline
     Params          &  \textbf{745.94}M    & 950.82M   &  963.44M &  1077.85M \\
     \hline
     FLOPs           &  \textbf{1337.59}G      &   5766.21G     &  5929.65G   & 6893.14G \\  
    \hline
  \end{tabular}
  }
  \caption{Computational cost comparison between ZePT and current OVSS methods. The FLOPs is computed based on input with spatial size $96\times 96 \times 96$ on the same single A100 GPU. 
  }
  \label{tab:Efficiency}
\end{table}

\begin{table}
  \setlength{\tabcolsep}{13pt}
  \centering
  \resizebox{\linewidth}{!}
  {
  \begin{tabular}{@{}l|c|c|c@{}}
    \hline
    Method & BTCV & LiTS & KiTS\\
    \hline
     nnUNet~\cite{isensee2021nnu}       & 82.23$\pm$2.07 & 77.15$\pm$3.47   & 85.18$\pm$1.26  \\
     Swin UNETR~\cite{tang2022self}     & 82.26$\pm$2.02      & 76.79$\pm$3.52      & 85.52$\pm$1.13 \\
     Universal~\cite{liu2023clip}       & 86.38$\pm$1.61  & 80.58$\pm$3.03 & 87.05$\pm$1.04 \\
    \hline
    ZePT       & \textbf{87.09$\pm$1.54}   & \textbf{81.66$\pm$2.79}  & \textbf{87.73$\pm$0.99} \\
    \hline
  \end{tabular}
  }
  \caption{5-fold cross-validation results on the BTCV~\cite{landman2015miccai}, LiTS~\cite{bilic2023liver}, and KiTS~\cite{heller2020international} validation dataset. We report the average DSC of 13 organs in BTCV, the Dice per case score of liver tumors in LiTS, and the DSC of kidney tumors in KiTS. 
  }
  \vspace{-10pt}
  \label{tab:so}
\end{table}

\noindent \textbf{Segmentation of Seen Organs and Tumors.}
As shown in~\cref{tab:so}, the segmentation performance of ZePT on seen organs surpasses strong baseline nnUNet~\cite{isensee2021nnu} and Swin UNETR~\cite{tang2022self} by at least absolute $4.83\%$ in DSC on BTCV~\cite{landman2015miccai}, $4.51\%$ in DSC for liver tumors on LiTS~\cite{bilic2023liver}, and $2.21\%$ in DSC for kidney tumors on KiTS~\cite{heller2020international}. Notably, ZePT achieves comparable or even better segmentation performance compared with the Universal model~\cite{liu2023clip}, which utilizes an assembly of 14 public datasets with a total of $3,410$ CT scans for training. The Universal model~\cite{liu2023clip} adopts a CLIP text encoder, which processes the organ and tumor names into text embeddings, to introduce the semantic relationship between anatomical structures. ZePT takes one step further by designing a more advanced architecture for visual feature extraction and incorporating additional medical domain knowledge to 
polish feature representations with diverse and fine-grained cues. 
These improvements demonstrate that ZePT can also segment seen organs and tumors with high accuracy.

\noindent \textbf{Qualitative Analysis.} \cref{fig:QR} shows the qualitative results and demonstrates the merits of ZePT. Most competing methods suffer from segmentation target incompleteness-related failures and misclassification of background regions as tumors (false positives).
ZePT produces sharper boundaries and generates results that are more consistent with the ground truth in comparison with all other models. 

\begin{figure}[htbp]
  \centering
  \includegraphics[width=\linewidth]{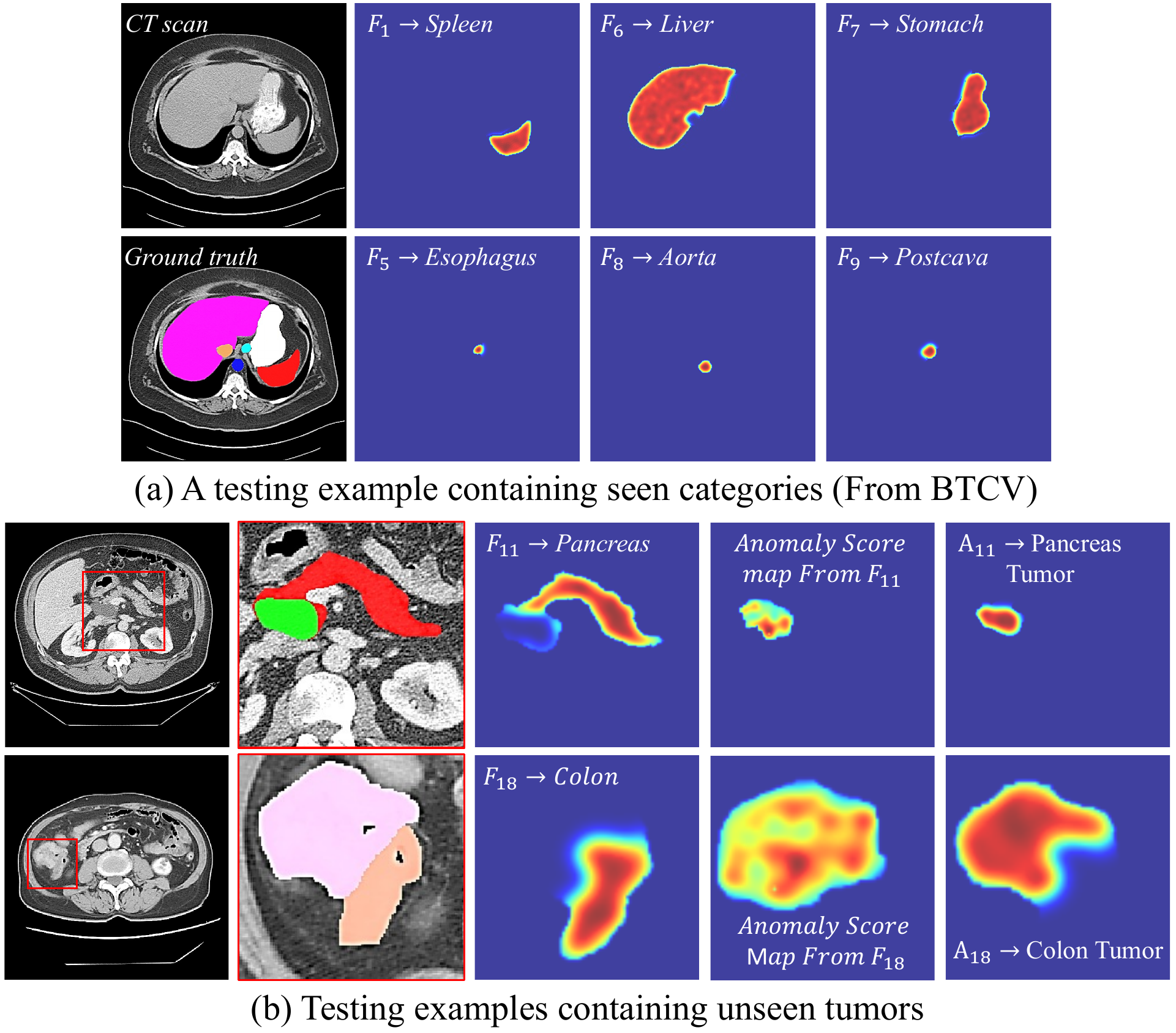}
  \caption{Visualization of query response maps. (a) A test sample containing seen categories from BTCV~\cite{landman2015miccai} evaluation set. (b) Two test samples, one from the MSD's pancreas tumor task~\cite{antonelli2022medical} and the other from the real-world colon tumor segmentation dataset. We can observe the query distribution on the different organs and tumors with obvious separation.
  The clear boundaries and high responses show the advantages of encouraging discriminative and disentangled queries to represent different objects, which benefits the segmentation of both seen and unseen categories.}
    \vspace{-10pt}
  \label{fig:ATTvis}
\end{figure}

We also visualize the query response maps of different seen organs and unseen tumors, as well as the anomaly score maps to illustrate the working mechanism of fundamental queries and advanced queries in ZePT. As shown in~\cref{fig:ATTvis} (a),
the different organ regions are confidently activated by fundamental queries $F_i$ ($F_1$ for spleen, $F_5$ for esophagus, $F_6$ for liver, \textit{etc}.). This advantage is attributed to the object-aware feature grouping, which enables each fundamental query to represent a corresponding organ.

In Row $1$ of~\cref{fig:ATTvis} (b), we can observe the pancreas region is captured by the fundamental query $F_{11}$, and the anomaly score maps derived from $F_{11}$ maintains high responses around the pancreas tumor region to indicate our model the potential location of unseen tumors. Furthermore, the advanced queries $A_{11}$ capture the region corresponding to the pancreas tumor according to the guidance of mask prompts derived from anomaly score maps. The same phenomenon can also be observed in the example of colon tumor segmentation in Row $2$ of~\cref{fig:ATTvis} (b). 
These visual examples fulfill the motivation of the self-prompting strategy that uses the anomaly score maps derived from fundamental queries as visual prompts to guide the advanced queries to segment unseen tumors. 

\subsection{Ablation Study and Discussions}



\noindent \textbf{Significance of Object-Aware Feature Grouping.}
We remove the object-aware feature grouping (OFG) and use the vanilla MaskFormer~\cite{cheng2022masked} to update the fundamental queries. We denote this variant as ZePT (w/o OFG) in~\cref{tab:NC}, which is observed to have a performance drop on both seen and unseen categories. This shows that OFG is crucial in learning object queries for region-level information.

\noindent \textbf{Importance of Query Refinement Decoder.} We replace the Query Refinement Decoder (QRD) with a vanilla transformer decoder to directly update all object queries together. It should be noted that this operation also eliminates the query-disentangling scheme since object queries are entangled into a single set and directly matched across all categories. We denote this variant as ZePT (w/o QRD) in~\cref{tab:NC}, which is observed to have a significant performance drop of $11.89\%$ in AUROC and $10.39\%$ in DSC for unseen colon tumor segmentation. This confirms the efficacy of QRD, designed to use visual prompts for updating advanced queries and to introduce interactions between the two sets of queries. This design enables both query sets to modify their representations for enhanced performance.

\noindent \textbf{Effectiveness of Prompt-Based Masked Attention.} As illustrated in~\cref{tab:NC}, we replace the prompt-based masked attention with a vanilla self-attention layer and directly concatenate the visual prompts (anomaly score maps) with image features to update the advanced queries. We denote this variant as ZePT (w/o PMA). 
This variant leads to decreased performance on both seen and unseen tumors, verifying that prompt-based masked attention is an effective way to leverage the visual prompts.

\noindent \textbf{Efficacy of the Query-Knowledge Alignment.} We remove the query-knowledge alignment and employ the CLIP image encoder to classify mask proposals in the same manner as conventional OVSS methods. We denote this variant as ZePT (w/o QKA) in~\cref{tab:NC}, which is observed to have a performance drop of $10.13\%$ in AUROC and $5.19\%$ in DSC on unseen colon tumors. Compared with an additional tuned CLIP image encoder, our feature-level cross-modal alignment between queries and domain knowledge directly and explicitly introduces high-level linguistic semantics into the visual representation, which is more effective and efficient.

\begin{table}
  \Huge
  \centering
  \resizebox{\linewidth}{!}
  {
  \begin{tabular}{@{}l|cc|ccc@{}}
    \toprule
    \multicolumn{1}{l|}{\multirow{2}{*}{Method}} & \multicolumn{2}{c|}{\makecell[c]{Liver Tumor \\in LiTS (Seen)}} & \multicolumn{3}{c}{\makecell[c]{Real-World Colon\\Tumor (Unseen)}}\\
    \cline{2-3}\cline{4-6}
    & AUROC$\uparrow$  & DSC$\uparrow$ & AUROC$\uparrow$ & FPR$_{95}$$\downarrow$ & DSC$\uparrow$  \\
    \hline
     ZePT (w/o OFG)      & 95.96$\pm$0.85                    & 80.84$\pm$3.17
     & 75.39      & 48.24         &31.28   \\
     ZePT (w/o QRD)      & 94.53$\pm$1.75                    & 80.11$\pm$3.98  
     & 72.46        & 58.93      & 25.84         \\ 
     ZePT (w/o PMA)      & 96.25$\pm$0.62                  & 81.00$\pm$3.01
     & 78.01 & 44.35  & 33.06\\
     ZePT (w/o QKA)      & 95.04$\pm$1.09                    & 80.61$\pm$3.34 
     & 74.22                   & 50.31             &31.04\\
     ZePT                & \textbf{96.82$\pm$0.47}  & \textbf{81.66$\pm$2.79}  
     & \textbf{84.35}  &\textbf{38.29}    & \textbf{36.23}\\
    \bottomrule
  \end{tabular}
  }
  \caption{Ablation study of different network components on LiTS and the real-world colon tumor segmentation dataset.}
  \label{tab:NC}
\end{table}


\noindent \textbf{Investigation on Self-Generated Visual Prompts.} 
Recently, many methods based on SAM~\cite{kirillov2023segment} leverage visual prompts for medical image segmentation~\cite{ma2023segment,wu2023medical,chen2023ma}. ZePT, using self-generated prompts adaptively derived from the embedding space, consistently outperforms these SAM-based methods, which adopt bounding boxes or points provided by the users as prompts (\cref{tab:selfprompted}). Moreover, our prompt-based masked attention can also handle box prompts. We observe that self-generated visual prompts can match or even surpass the performance of strong manual prompts, like relaxed boxes, a finding corroborated by our visualization of anomaly score maps in~\cref{fig:ATTvis} (b). Such flexible and adaptive self-generated prompts are crucial for unseen tumor segmentation scenarios, where acquiring even bounding box prompts is challenging.
\begin{table}
  \centering
  \resizebox{0.85\linewidth}{!}
  {
  \begin{tabular}{@{}l|cc@{}}
    \hline
    Methods with Different Prompts
     & DSC$\uparrow$ & NSD$\uparrow$\\
    \hline
     MedSAM~\cite{ma2023segment} (relaxed 3d bbx prompt)  & 26.18             & 36.25 \\
     MSA~\cite{wu2023medical} (1 point prompt)          & 27.88             & 39.06 \\
     MA-SAM~\cite{chen2023ma} (relaxed 3d bbx prompt)     & 29.39             & 41.11 \\
     ZePT (relaxed 3d bbx prompt)             & 35.89    & 48.04 \\
     ZePT                                   & \textbf{36.23}             & 48.78  \\
    \hline
  \end{tabular}
  }
  \caption{Comparisons between ZePT and SAM-based~\cite{kirillov2023segment} medical image segmentation methods~\cite{ma2023segment,wu2023medical,chen2023ma} on real-world colon tumor segmentation dataset. We report DSC and Normalized Surface Distance (NSD).}
  \vspace{-10pt}
  \label{tab:selfprompted}
\end{table}

\noindent \textbf{Discussions about Limitations.} The zero-shot segmentation performance of ZePT on unseen tumors still falls behind supervised models fully trained on these tumors. Although the comparison is unfair, it indicates there is still much room for improvement. We hope our work can shed some light on designing models with zero-shot abilities for medical imaging tasks.
\section{Conclusions}
\label{conclusion}
In this work, we propose ZePT, a novel framework based on query-disentangling and self-prompting for zero-shot pan-tumor segmentation. 
We disentangle the object queries into two subsets and decouple their learning process into two stages. ZePT exploits discriminative and object-level feature representation for organs and tumors. We introduce a self-prompting strategy to adaptively localize abnormalities for guiding the queries to be aware of the pathological changes among visual contexts. Additionally, we perform query-knowledge alignment at the feature level to further enhance the model's generalization capabilities. The significant performance improvements of ZePT on various organ and tumor segmentation tasks validate its effectiveness.
\par

{
    \small
    \bibliographystyle{ieeenat_fullname}
    \bibliography{main}
}

\clearpage
\setcounter{page}{1}
\maketitlesupplementary

\section{Appendix}
\label{sec:Appendix}
\subsection{Dataset Details}
\noindent \textbf{Training:} In Stage-I, we assemble the training sets of $8$ public datasets, including Pancreas-CT~\cite{roth2015deeporgan}, AbdomenCT-1K~\cite{ma2021abdomenct}, CT-ORG~\cite{rister2020ct}, CHAOS~\cite{kavur2021chaos}, 
AMOS22~\cite{ji2022amos}, BTCV~\cite{landman2015miccai}, WORD~\cite{luo2022word} and TotalSegmentator~\cite{wasserthal2022totalsegmentator}.
These datasets exclusively contained organ labels. In {Stage-II}, we add CT images from the training sets of LiTS~\cite{bilic2023liver} and KiTS~\cite{heller2020international}. The overall seen categories used for training consist of $25$ organ classes and $2$ tumor classes (Liver Tumor and Kidney Tumor). 

(1) Pancreas-CT~\cite{roth2015deeporgan} consists of $82$ contrast-enhanced abdominal CT volumes. This dataset only provides the pancreas
label annotated by an experienced radiologist, and all CT scans have no pancreatic tumor.

(2) AbdomenCT-1K~\cite{ma2021abdomenct} consists of $1112$ CT scans from five datasets and includes annotations for the liver, kidney, spleen, and pancreas.

(3) CT-ORG~\cite{rister2020ct} comprises $140$ CT images containing 6 organ classes. This dataset is sourced from eight different medical centers. Predominantly, these images display liver lesions, encompassing both benign and malignant types.

(4) CHAOS~\cite{kavur2021chaos} provides $40$ CT scans including healthy abdomen organs without any pathological abnormalities (tumors, metastasis, and so on) for multi-organ segmentation. 

(5) AMOS22~\cite{ji2022amos}, the multi-modality abdominal multi-organ segmentation challenge of $2022$, contains $500$ CT scans with voxel-level annotations of $15$ abdominal organs.

(6) BTCV dataset~\cite{landman2015miccai} contains 30 subjects of abdominal CT scans where 13 organs are annotated by interpreters under the supervision of radiologists at Vanderbilt University Medical Center.

(7) WORD~\cite{luo2022word} collects 150 CT scans from 150 patients before the radiation therapy in a single center. All of them are scanned by a SIEMENS CT scanner without appearance enhancement. Each CT volume consists of $159$ to $330$ slices of $512 \times 512$ pixels. All scans of WORD dataset are exhaustively annotated with 16 anatomical organs.

(8) TotalSegmentator~\cite{wasserthal2022totalsegmentator} consists of $1024$ CT scans of different body parts with a total of $104$ labeled anatomical structures. Only organ labels are adopted in this paper.

(9) LiTS~\cite{bilic2023liver} contains $131$ and $70$ contrast-enhanced abdominal CT scans for training and testing, respectively. The
data set was acquired by different scanners and protocols at six different clinical sites, with a largely varying in-plane resolution from $0.55$ to $1.0$ mm and slice spacing from $0.45$ to $6.0$ mm.

(10) KiTS~\cite{heller2020international} includes $210$ training cases and $90$ testing cases with annotations provided by the University of Minnesota Medical Center. Each CT scan has one or more kidney tumors.

\noindent \textbf{Inference:} 
We employ the MSD dataset~\cite{antonelli2022medical} that encompasses a range of segmentation tasks for five tumor types in CTs. Among these, pancreas tumors, lung tumors, colon tumors, and hepatic vessel tumors belong to unseen categories. 
A real-world, private dataset containing $388$ 3D CT volumes of four distinct colon tumor subtypes is also utilized for testing.

(1) MSD CT Tasks~\cite{antonelli2022medical} includes liver, lung, pancreas, colon, hepatic vessel, and spleen tasks for a total of $947$ CT scans with $4$ organs and $5$ tumors.

(2) To further evaluate the proposed method, we collect a large real-world colon cancer CT dataset, which consists of 388 patients diagnosed with colon cancer. For each patient, an abdominal CT (venous phase) scan is collected, and the tumor region is annotated by an experienced gastroenterologist and later verified by another senior radiologist. During the annotation phase, the physicians are also provided with the corresponding post-surgery pathological report to narrow down the search area for the tumors. All the scans share the same in-plane dimension of $512 \times 512$, and the dimension along the z-axis ranges from $36$ to $146$, with a median of $91$. The in-plane spacing ranges from $0.60\times0.60$ to $0.98\times0.98$ mm, with a median of $0.76 \times 0.76$ mm, and the z-axis spacing is from $5.0$ to $7.5$ mm, with a median of $5.0$ mm. There are four tumor subtypes in this dataset. The full name and incidence count for each disease are shown in~\cref{tab:CRCdataset}. It is important to note that signet ring cell adenocarcinoma and adenosquamous carcinoma constitute only an exceedingly small proportion of all colon cancer cases, illustrating the long-tailed distribution characteristic of real-world disease incidence.

\begin{table}
  \small
  \centering
  \resizebox{0.8\linewidth}{!}
  {
  \begin{tabular}{@{}l|c@{}}
    \hline
    Full Name of Tumor Subtype & Count
    \\
    \hline
     Adenocarcinoma   & 278  \\
     Mucinous adenocarcinoma  & 64 \\
     Signet ring cell adenocarcinoma (rare) & 29\\
     Adenosquamous carcinoma (rare)  & 17  \\
    \hline
  \end{tabular}
  }
  \caption{Dataset details of the real-world colon tumor segmentation dataset.}
  \label{tab:CRCdataset}
\end{table}

%
%
%

\begin{table*}
  \Huge
  \centering
  \resizebox{\linewidth}{!}
  {
  \begin{tabular}{@{}l|c|c|l@{}}
    \hline
    Datasets & \#Target & \#Scans & Annotated categories
    \\
    \hline
     Pancreas-CT~\cite{roth2015deeporgan}   & 1 & 82 & Pancreas  \\
     AbdomenCT-1K~\cite{ma2021abdomenct}    & 4 & 1000 & Spleen, Kidney, Liver, Pancreas \\
     CT-ORG~\cite{rister2020ct}             & 4 & 140 &  Lung, Liver, Kidneys and Bladder\\
     CHAOS~\cite{kavur2021chaos}            & 4 & 40 &  Liver, Left Kidney, Right Kidney, Spleen\\
     AMOS22~\cite{ji2022amos}               & 15 & 500 & Spl, RKid, LKid, Gall, Eso, Liv, Sto, Aor, IVC, Pan, RAG, LAG, Duo, Bla, Pro/UTE\\
     BTCV~\cite{landman2015miccai}          & 13 & 30 & Spl, RKid, LKid, Gall, Eso, Liv, Sto, Aor, IVC, R\&SVeins, Pan, RAG, LAG\\
     WORD~\cite{luo2022word}                & 16 & 150 & Spl, RKid, LKid, Gall, Eso, Liv, Sto, Pan, RAG, Duo, Col, Int, Rec, Bla, LFH, RFH  \\
     TotalSegmentator~\cite{wasserthal2022totalsegmentator}  & 104 & 1024 & Spl, RKid, LKid, Gall, Liv, Sto, Pan, RAG, LAG, Eso, Duo, Small Bowel, Colon, and so on. \\
     LiTS~\cite{bilic2023liver}             & 2 & 201 & Liver, \textit{Liver Tumor}  \\
     KiTS~\cite{heller2020international}    & 2 & 300 & Kidney, \textit{Kidney Tumor}  \\
     \hline
     MSD CT Tasks~\cite{antonelli2022medical}   & 9 & 947 & \makecell[l]{Spl, Liver, \textit{Liver Tumor}, \textit{Lung Tumor}, \textit{Colon Tumor}, Pancreas and \textit{Pancreas Tumor}, \\ Hepatic Vessel and \textit{Hepatic Vessel Tumor}}  \\
     \cline{4-4}
     real-world colon tumor dataset             & 4 & 388 & \textit{Colon Tumor} with four subtypes  \\
    \hline
  \end{tabular}
  }
  \caption{The information for all datasets used for training and testing.}
  \label{tab:SDATA}
\end{table*}

We summarize all the datasets in~\cref{tab:SDATA}. As our main objective is not dealing with partial label problem, we directly adopt the successful data processing strategy in~\cite{liu2023clip}. Concretely, we pre-process CT scans using isotropic spacing and uniformed intensity scale to reduce the domain gap among various datasets. Then we unify the label index for all datasets. For these datasets (KiTS, WORD, AbdomenCT-1K, and CT-ORG), which do not distinguish between the left and right organs, we split the organ (Kidney, Adrenal Gland, and Lung) into left part and right part. Since we formulate each organ segmentation result as a binary mask, we can organize the segmentation ground truth for these overlapped organs independently in a binary mask manner. During traning, we associate fundamental queries with organ classes, and advanced queries with tumor classes. The corresponding relationship is shown in~\cref{tab:labelindex}.

\begin{table*}
  \centering
  \resizebox{\linewidth}{!}
  {
  \begin{tabular}{@{}l|c|l|c@{}}
    \hline
    Fundamental Queries $\rightarrow$ Organ Categories & Organ Label Index  & Advanced Queries $\rightarrow$ Tumor Categories & Tumor Label Index
    \\
    \hline
     $F_1 \rightarrow$  Spleen      &1                              & $A_1 \rightarrow$ Spleen Tumor (unseen)  & 26\\
     $F_2 \rightarrow$ Right Kidney &2                              & $A_2 \rightarrow$ Kidney Tumor (seen) & 27\\
     $F_3 \rightarrow$ Left Kidney   &3                             & $A_3 \rightarrow$ Kidney Cyst (unseen)  & 28\\
     $F_4 \rightarrow$ Gall Bladder  &4                             & $A_4 \rightarrow$ Gall Bladder Tumor (unseen) & 29 \\
     $F_5 \rightarrow$ Esophagus   &5                               & $A_5 \rightarrow$ Esophagus Tumor (unseen) &30\\
     $F_6 \rightarrow$ Liver    &6                                  & $A_6 \rightarrow$ Liver Tumor (seen)  &31\\
     $F_7 \rightarrow$ Stomach  &7                                  & $A_7 \rightarrow$ Stomach Tumor (unseen) &32\\
     $F_8 \rightarrow$ Aorta    &8                                  & $A_8 \rightarrow$ Aortic Tumor (unseen) &33\\
     $F_9 \rightarrow$ Postcava  &9                                 & $A_9 \rightarrow$ Postcava Tumor Thrombus (unseen)  &34\\
     $F_{10} \rightarrow$ Portal Vein and Splenic Vein  & 10        & $A_{10} \rightarrow$ Portal Vein Tumor Thrombus (unseen) & 35\\
     $F_{11} \rightarrow$ Pancreas  & 11                            & $A_{11} \rightarrow$ Pancreas Tumor (unseen) & 36\\
     $F_{12} \rightarrow$ Right Adrenal Gland  & 12                 & $A_{12} \rightarrow$ Adrenal Tumor (unseen) & 37\\
     $F_{13} \rightarrow$ Left Adrenal Gland  & 13                  & $A_{13} \rightarrow$ Adrenal Cyst (unseen) & 38\\
     $F_{14} \rightarrow$ Duodenum & 14                             & $A_{14} \rightarrow$ Duodenal Tumor (unseen) & 39\\
     $F_{15} \rightarrow$ Hepatic Vessel & 15                       & $A_{15} \rightarrow$ Hepatic Vessel Tumor (unseen) & 40\\
     $F_{16} \rightarrow$ Right Lung & 16                           & $A_{16} \rightarrow$ Lung Tumor (unseen) & 41\\
     $F_{17} \rightarrow$ Left Lung & 17                            & $A_{17} \rightarrow$ Lung Cyst (unseen) & 42\\
     $F_{18} \rightarrow$ Colon & 18                                & $A_{18} \rightarrow$ Colon Tumor (unseen) & 43\\
     $F_{19} \rightarrow$ Intestine & 19                            & $A_{19} \rightarrow$ Small Intestinal Neoplasm (unseen) & 44\\
     $F_{20} \rightarrow$ Rectum & 20                               & $A_{20} \rightarrow$  Rectal Tumor (unseen) & 45\\
     $F_{21} \rightarrow$ Bladder & 21                              &  & \\
     $F_{22} \rightarrow$ Prostate & 22                             &  & \\
     $F_{23} \rightarrow$ Left Head of Femur & 23                   &  & \\
     $F_{24} \rightarrow$ Right Head of Femur & 24                  &  & \\
     $F_{25} \rightarrow$ Celiac Trunk & 25                         &  & \\
     
    \hline
  \end{tabular}
  }
  \caption{The correspondence between object queries and the categories they are responsible for. Only organ categories and seen tumor categories involve voxel-wise annotations. Queries tasked with identifying and segmenting seen organs and tumors receive supervision from both ground truth mask annotations and query-knowledge alignment. Advanced queries responsible for identifying and segmenting unseen tumor categories only have weak supervision from query-knowledge alignment.}
  \label{tab:labelindex}
\end{table*}

\subsection{Qualitative Analysis on Real-World Colon Tumor Segmentation Dataset.}
\begin{figure*}
  \centering
  \includegraphics[width=\linewidth]{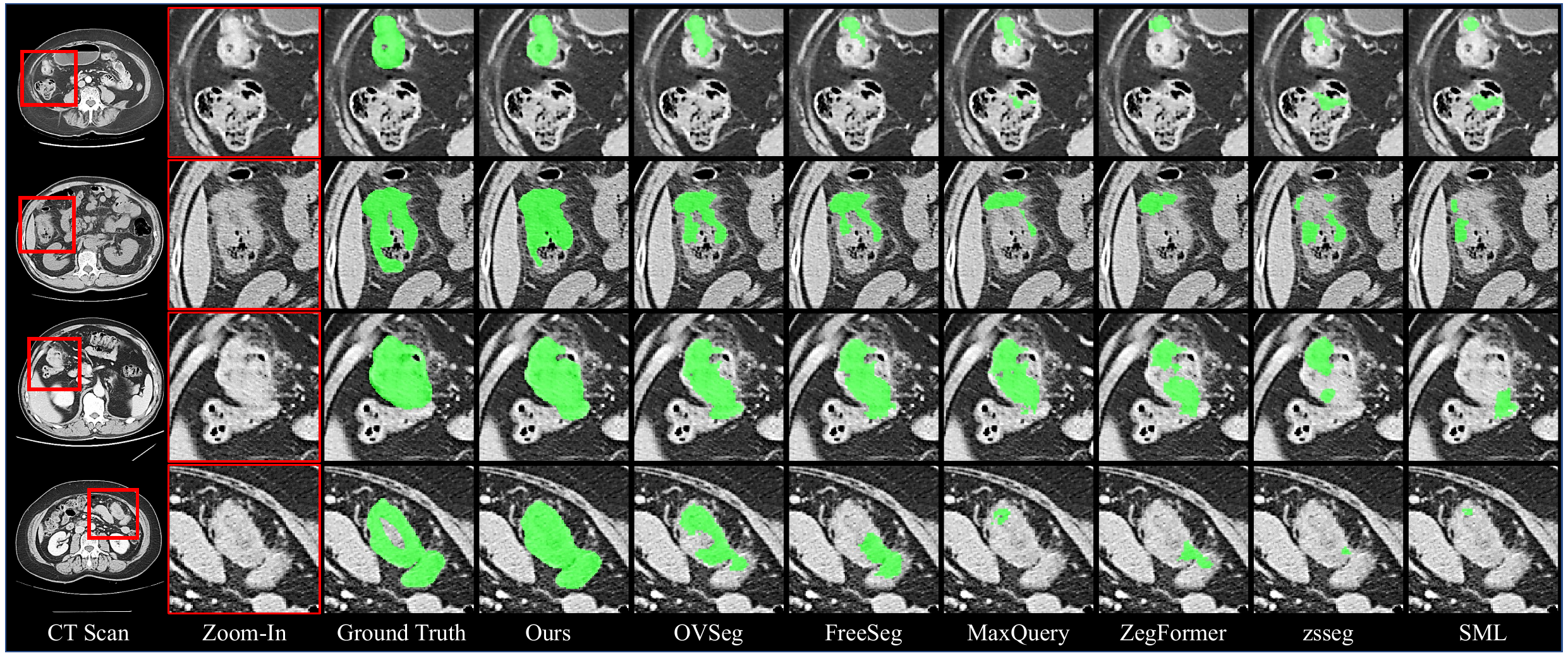}
  \caption{Qualitative visualizations on real-world colon tumor segmentation dataset. We compare ZePT with other advanced OVSS methods and OOD detection methods in a zero-shot manner.}
  \label{fig:CRCQR}
\end{figure*}

For qualitative analysis on real-world colon tumors, we present visualizations of segmentation results in~\cref{fig:CRCQR}. This shows that our approach achieves much better zero-shot segmentation performance on real-world colon tumors compared with other methods.

\subsection{Detailed Results of Real-World Colon Tumor Segmentation Analysis.}
In~\cref{tab:DRCRC}, we provide a detailed analysis of the detection and segmentation performance of ZePT and OVSeg~\cite{liang2023open}, the second-ranked method, for four subtypes of colon tumors within the real-world colon tumor dataset. 
ZePT consistently demonstrates superior detection and segmentation capabilities over OVSeg for both commonly encountered and rare subtypes of colon tumors. In the analysis of segmentation performance for the common colon cancer subtypes, Adenocarcinoma and Mucinous Adenocarcinoma, ZePT exhibited a substantial enhancement over OVSeg~\cite{liang2023open} in terms of DSC, achieving increases of $33.80\%$ and $22.66\%$, respectively. In the segmentation of rare colon cancer subtypes, namely Signet Ring Cell Adenocarcinoma and Adenosquamous Carcinoma, ZePT also significantly outperformed OVSeg, achieving DSC improvements of $19.12\%$ and $5.14\%$, respectively. These results highlight ZePT's superior performance and its promising ability for zero-shot tumor segmentation in real-world settings. Since these rare disease types are individually infrequent, it is impossible to collect them completely. Therefore, we address the thorny problem by exploring and enhancing the model's zero-shot segmentation capability.

\begin{table*}[htbp]
  \Huge
  \centering
  \resizebox{\linewidth}{!}
  {
  \begin{tabular}{@{}l|ccc|ccc|ccc|ccc|ccc@{}}
    \toprule
    \multicolumn{1}{l}{\multirow{2}{*}{Method}} & \multicolumn{3}{|c|}{Adenocarcinoma} & \multicolumn{3}{c|}{Mucinous adenocarcinoma} & \multicolumn{3}{c|}{Signet ring cell adenocarcinoma} & \multicolumn{3}{c|}{Adenosquamous carcinoma} & \multicolumn{3}{c}{Average}\\
    \cline{2-4}\cline{5-7}\cline{8-10} \cline{11-13} \cline{14-16}     
     & AUROC$\uparrow$ & FPR$_{95}$$\downarrow$ & DSC$\uparrow$ & AUROC$\uparrow$ & FPR$_{95}$$\downarrow$ & DSC$\uparrow$
    & AUROC$\uparrow$ & FPR$_{95}$$\downarrow$ & DSC$\uparrow$ & AUROC$\uparrow$ & FPR$_{95}$$\downarrow$ & DSC$\uparrow$
    & AUROC$\uparrow$ & FPR$_{95}$$\downarrow$ & DSC$\uparrow$\\
    \hline
     OVSeg~\cite{liang2023open}            & 70.06 & 63.71 & 16.51         & 70.01 & 63.89 & 16.44         & 69.93 & 65.73 & 15.66             & 69.80 & 66.03 & 15.59             & 69.95 & 64.84 & 16.05 \\
    \hline
    ZePT    & \textbf{96.44} & \textbf{18.58} & \textbf{50.31} 
    & \textbf{88.29} & \textbf{31.64} & \textbf{39.10}
    & \textbf{81.80}   & \textbf{41.05}  & \textbf{34.78} 
    & \textbf{70.87} & \textbf{61.89} & \textbf{20.73}  
    & \textbf{84.35}  &\textbf{38.29}    & \textbf{36.23} \\
    \bottomrule
  \end{tabular}
  }
  \caption{Detection and segmentation performance of four colon tumor subtypes on real-world colon tumor dataset. We compare ZePT with the second-ranked method OVSeg~\cite{liang2023open}. ZePT markedly surpasses OVSeg in terms of detection and segmentation efficacy for both prevalent and rare types of colon tumors.
  }
  \label{tab:DRCRC}
\end{table*}

\subsection{Additional Ablation Experiments}
\subsubsection{Different Choices of Text Encoder.}
We evaluate the performance disparities arising from the use of various models as text encoders for generating text embeddings. In~\cref{tab:TEXTSEQ}, we observe that employing ClinicalBERT~\cite{alsentzer2019publicly} as the text encoder yields a slightly higher Dice Similarity Coefficient (DSC), with improvement of $0.39\%$ compared to the use of the CLIP Text Encoder~\cite{radford2021learning} in the context of the real-world colon tumor dataset. Therefore, we use the ClinicalBERT~\cite{alsentzer2019publicly} as the default text encoder setting for ZePT.

\subsubsection{Effectiveness of Medical Domain Knowledge.}
We examine the performance disparities in generating text embeddings for each organ and tumor when using a conventional prompt~\cite{liu2023clip} (\textit{e.g.}, ``a computerized tomography of a [CLS]'') versus employing additional domain knowledge automatically derived from the Large Language Model GPT4~\cite{openai2023gpt4}. As shown in~\cref{tab:TEXTSEQ}, incorporating additional domain knowledge results in a notable enhancement in text embedding efficacy, evidenced by an increase of $1.54\%$ in the DSC when using ClinicalBERT~\cite{alsentzer2019publicly} as the text encoder, and a $1.51\%$ increase when adopting CLIP Text Encoder~\cite{radford2021learning}, in the context of the real-world colon tumor dataset. Incorporating medical domain knowledge into the model enhances it with advanced high-level information and detailed visual cues. This improvement boosts the model's discriminative and generalization capabilities. Furthermore, the medical domain knowledge, which is initially auto-generated and then refined by medical professionals, will also be made publicly available alongside the source code.

\begin{table}
  \Huge
  \centering
  \resizebox{\linewidth}{!}
  {
  \begin{tabular}{@{}l|l|c@{}}
    \hline
    Text Encoder & text sequence & DSC$\uparrow$ \\
    \hline
    CLIP Text Encoder~\cite{radford2021learning} & A computerized tomography of a [CLS].  &   34.33 \\
    CLIP Text Encoder~\cite{radford2021learning}  & A computerized tomography of a [CLS]. + Knowledge  & 35.84   \\
    ClinicalBERT~\cite{alsentzer2019publicly}  & A computerized tomography of a [CLS].  & 34.69    \\
    ClinicalBERT~\cite{alsentzer2019publicly}  & A computerized tomography of a [CLS]. + Knowledge & \textbf{36.23}  \\
    \hline
  \end{tabular}
  }
  \caption{Ablation study of the additional medical domain knowledge on the real-world colon tumor segmentation dataset. We also compare the performance disparities in adopting different models as the text encoder (CLIP text encoder vs. ClinicalBERT). 
  }
  \label{tab:TEXTSEQ}
\end{table}

\subsubsection{Importance of Pretraining: One-Stage vs. Two-Stage.}
In ZePT, we initially pretrain fundamental queries on datasets exclusively containing organ labels to achieve multi-organ segmentation in Stage-I, subsequently fine-tuning these queries and training advanced queries in Stage-II. However, it is also feasible to bypass Stage-I entirely and directly train the whole model following the training protocol of Stage-II. The performance disparities between one-stage and two-stage training approaches for ZePT are summarized in~\cref{tab:onevstwo}. ZePT, when trained in two stages, significantly outperforms its one-stage counterpart, demonstrating improved zero-shot colon tumor segmentation. The performance metrics show an absolute increase of at least $5.87\%$ in AUROC, $5.74\%$ in FPR$_{95}$, and $3.04\%$ in DSC. Furthermore, we observed that the one-stage ZePT variant exhibits significant instability in the initial training phases and necessitates an extended number of epochs for convergence. These phenomena are primarily due to the omission of the initial pre-training stage for fundamental queries, which results in the model's inadequate understanding of anatomical structures, such as organs. As a result, the visual prompts derived from fundamental queries are ineffective in capturing essential information. This inefficacy hinders advanced queries, dependent on the visual prompts, from learning significant features, leading to a compromised feature representation that impairs the model's overall performance. The experimental findings highlight the necessity of a two-stage training approach for the model and reinforce our design insight of commencing with fundamental query training followed by its application in guiding the training of advanced queries.

\begin{table}
  \small
  \centering
  \resizebox{\linewidth}{!}
  {
  \begin{tabular}{@{}l|ccc@{}}
    \hline
    \multicolumn{1}{l|}{\multirow{2}{*}{Method}} & \multicolumn{3}{c}{Real-World Colon Tumor dataset}\\
    \cline{2-4}
    & AUROC$\uparrow$ & FPR$_{95}$$\downarrow$ & DSC$\uparrow$  \\
    \hline
     ZePT-One Stage (No Pretraining)       & 78.48           & 44.03            &33.19\\
     ZePT-Two Stage (With Pretraining)     & \textbf{84.35}  &\textbf{38.29}    & \textbf{36.23}\\
    \hline
  \end{tabular}
  }
  \caption{Ablation study on the impact of pretraining fundamental queries for multi-organ segmentation in Stage-I.}
  \label{tab:onevstwo}
\end{table}

\subsubsection{Object-Aware Feature Grouping vs. other alternatives.}
The object-aware feature grouping (OFG) strategy enables object queries in ZePT to acquire organ-level semantics. We compare OFG with a close alternative method, GroupViT~\cite{xu2022groupvit}, which generates a set of queries as clustering centers and clusters pixels with similar semantics via $\operatorname{Gumbel-Softmax}$~\cite{jang2016categorical,maddison2016concrete} operation. As shown in~\cref{tab:abonOFG}, adopting OFG results in an improvement of $3.36\%$ in AUROC, $3.35\%$ in FPR$_{95}$, and $1.76\%$ in DSC, compared to the performance achieved with GroupViT~\cite{xu2022groupvit}. The results further confirm the efficacy of the OFG strategy. Conversely, the bottom-up clustering approach based on pixel semantics in GroupViT~\cite{xu2022groupvit} is unsuitable for the zero-shot tumor segmentation (ZSTS) task. This task necessitates differentiating the unseen tumor region from adjacent regions with similar semantics. Therefore, instead of using Gumbel-Softmax for bottom-up pixel grouping, our approach employs it to contrast visual features with specialized object queries. This method prevents the blending of target and adjacent disturbing regions.

\begin{table}
  \Huge
  \centering
  \resizebox{\linewidth}{!}
  {
  \begin{tabular}{@{}l|ccc@{}}
    \hline
    \multicolumn{1}{l|}{\multirow{2}{*}{Method}} & \multicolumn{3}{c}{Real-World Colon Tumor dataset}\\
    \cline{2-4}
    & AUROC$\uparrow$ & FPR$_{95}$$\downarrow$ & DSC$\uparrow$  \\
    \hline
     ZePT (GroupViT~\cite{xu2022groupvit} backbone)               & 80.99           & 41.64            & 34.47 \\
     ZePT (MaskFormer~\cite{cheng2022masked} backbone + OFG)      & \textbf{84.35}  &\textbf{38.29}    & \textbf{36.23}\\
    \hline
  \end{tabular}
  }
  \caption{Ablation study of the object-aware feature grouping (OFG) strategy and its alternative.}
  \label{tab:abonOFG}
\end{table}

\subsubsection{Using Different Training Data in Stage-II.}
During Stage-II of ZePT's training process, we utilize the LiTS~\cite{bilic2023liver} and KiTS~\cite{heller2020international} datasets, which include two tumor categories, \textit{i}.\textit{e}., liver and kidney tumors, respectively. We explore the impact of integrating additional tumor categories into Stage-II of ZePT's training process. We experiment with various tumor categories and corresponding datasets: liver tumor (LiTS~\cite{bilic2023liver}), kidney tumor (KiTS~\cite{heller2020international}), lung tumor (MSD lung task~\cite{antonelli2022medical}), and pancreas tumor (MSD pancreas task~\cite{antonelli2022medical}). The efficacy of ZePT, trained across these diverse tumor categories, are evaluated using the MSD hepatic vessel tumor task~\cite{antonelli2022medical} and the real-world colon tumor dataset for zero-shot tumor segmentation performance. The results are summarized in~\cref{tab:abondataset}. We observed several notable intrinsic phenomena. 

Firstly, training on images with liver tumors significantly enhances the zero-shot segmentation performance for Hepatic Vessel tumors, more so than training with other tumor categories. This improvement can be attributed to the visual similarity between liver and hepatic vessel tumors in imaging, which results in a substantially higher zero-shot segmentation performance for hepatic vessel tumors following exposure to liver tumors. 

Secondly, progressively increasing the number of CT scans and diversifying the tumor categories included in the training process leads to a stable and gradual improvement in the model's zero-shot segmentation performance on unseen tumors. 

Thirdly, the model demonstrates significantly enhanced zero-shot segmentation capabilities for unseen tumors that share similar imaging characteristics with the tumor types included in the training set. This is in stark contrast to its performance on tumors with imaging features distinctly different from those in the training set. This phenomenon accounts for the findings in~\cref{tab:MSD}, where the model shows superior zero-shot segmentation for hepatic vessel tumors compared to others, yet demonstrates relatively lower efficacy for lung and colon tumors.

\begin{table*}
  \Huge
  \centering
  \resizebox{\linewidth}{!}
  {
  \begin{tabular}{@{}l|c|ccc|ccc@{}}
    \hline
    \multicolumn{1}{l|}{\multirow{2}{*}{Tumor Categories in Dataset}} & \multirow{2}{*}{\#Scans} & \multicolumn{3}{c|}{MSD~\cite{antonelli2022medical} Hepatic Vessel Tumor} & \multicolumn{3}{c}{Real-World Colon Tumor dataset}\\
    \cline{3-5} \cline{6-8}
    & & AUROC$\uparrow$ & FPR$_{95}$$\downarrow$ & DSC$\uparrow$  & AUROC$\uparrow$ & FPR$_{95}$$\downarrow$ & DSC$\uparrow$  \\
    \hline
     (Liver Tumor, Kidney Tumor)                              & 341   & 91.57   & 20.64  & 52.94          & 84.35  &38.29   & 36.23\\
     (Lung Tumor, Kidney Tumor)                               & 274   & 86.80   & 33.31  & 40.57           & 82.56  &40.95   & 35.02\\
     (Lung Tumor, Liver Tumor)                                & 195   & 90.72   & 23.98  & 50.39           & 78.50  &44.01   & 33.21\\
     (Pancreas Tumor, Liver Tumor)                            & 413   & 92.58   & 19.40  & 53.83           & 86.82  &34.36   & 37.95\\
     (Pancreas Tumor, Liver Tumor, Kidney Tumor)              & 623   & 93.41   & 17.79  & 54.99           & 87.71  &32.29   & 38.87\\
     (Pancreas Tumor, Liver Tumor, Kidney Tumor, Lung Tumor)  & 687   & 94.26   & 17.05  & 55.76           & 89.04  &30.68   & 39.52\\
    \hline
  \end{tabular}
  }
  \caption{Ablation study of using different training data in Stage-II.}
  \label{tab:abondataset}
\end{table*}

\begin{figure}[htbp]
  \centering
  \includegraphics[width=\linewidth]{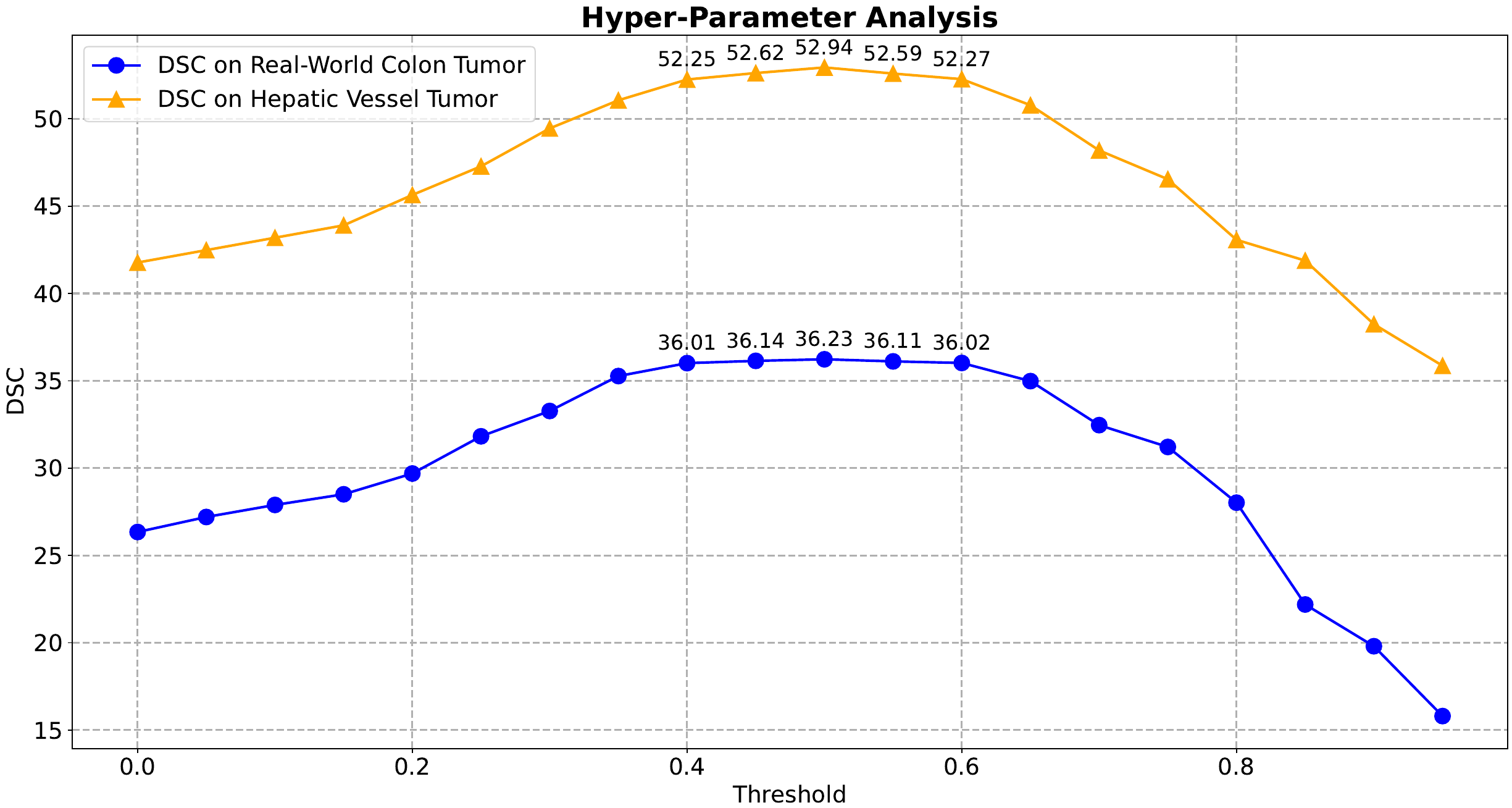}
  \caption{Hyper-Parameter Analysis of the threshold which determines the quality of mask prompts.}
  \label{fig:HPA}
\end{figure}

\subsubsection{Methodological clarification of Gumbel-Softmax.}
The main motivation of using Gumbel-Softmax is to make the $\operatorname{argmax}$ operation in eq.(4) differentiable~\cite{van2017neural}, where $\operatorname{argmax}$ enables the exclusive one-hot hard assignment of each query. This helps queries focus on distinct visual areas without overlap. We conduct experiments to examine the effectiveness of adopting one-hot hard assignment with Gumbel-Softmax. Our findings reveal that, in comparison to the straightforward use of cross-attention, our choice of incorporating one-hot hard assignment alongside Gumbel-Softmax enhances the DSC by $4.95\%$ on the real-world colon tumor dataset. Also, substituting Softmax with Gumbel-Softmax in equation (2) leads to a $2.86\%$ decrease in DSC on the real-world colon tumor dataset. We assume this occurs because the soft assignment by Softmax, prior to the hard assignment of local features, endows the queries with a global receptive field, allowing them to benefit from long-range contexts.

\subsection{Hyper-Parameter Analysis}
As described in~\cref{PUQ}, we derive anomaly score maps from the affinity between visual features and fundamental queries. Subsequently, these anomaly score maps undergo min-max normalization, followed by the application of a $0.5$ threshold to derive the mask prompts. Then the prompt-based masked attention enables advanced queries to focus on regions specified by the mask prompts, effectively controlling the receptive field of these queries. A critical hyper-parameter in this process, namely the threshold, is instrumental in determining the quality of mask prompts. We conduct experiments on the MSD hepatic vessel tumor task~\cite{antonelli2022medical} and the real-world colon tumor dataset to analyse the influence of the threshold in~\cref{fig:HPA}. We observed that the model achieves optimal performance in zero-shot tumor segmentation on two datasets when the threshold is set at $0.5$. Within the threshold range of $0.4$ to $0.6$, the model demonstrates high stability and robustness to threshold variations, with no significant changes in performance. However, outside this range, there is a marked decline in performance. This is due to the fact that a threshold near $0$ results in minimal masking of visual features, hindering advanced queries from focusing on key visual cues in the lesion area. In contrast, a threshold near $1$ leads to excessive masking of visual features, restricting the receptive field of advanced queries and limiting their access to sufficient effective information, consequently causing a marked decrease in performance. Therefore, we adopt 0.5 as the default setting for the threshold.

\begin{figure*}
  \centering
  \includegraphics[width=\linewidth]{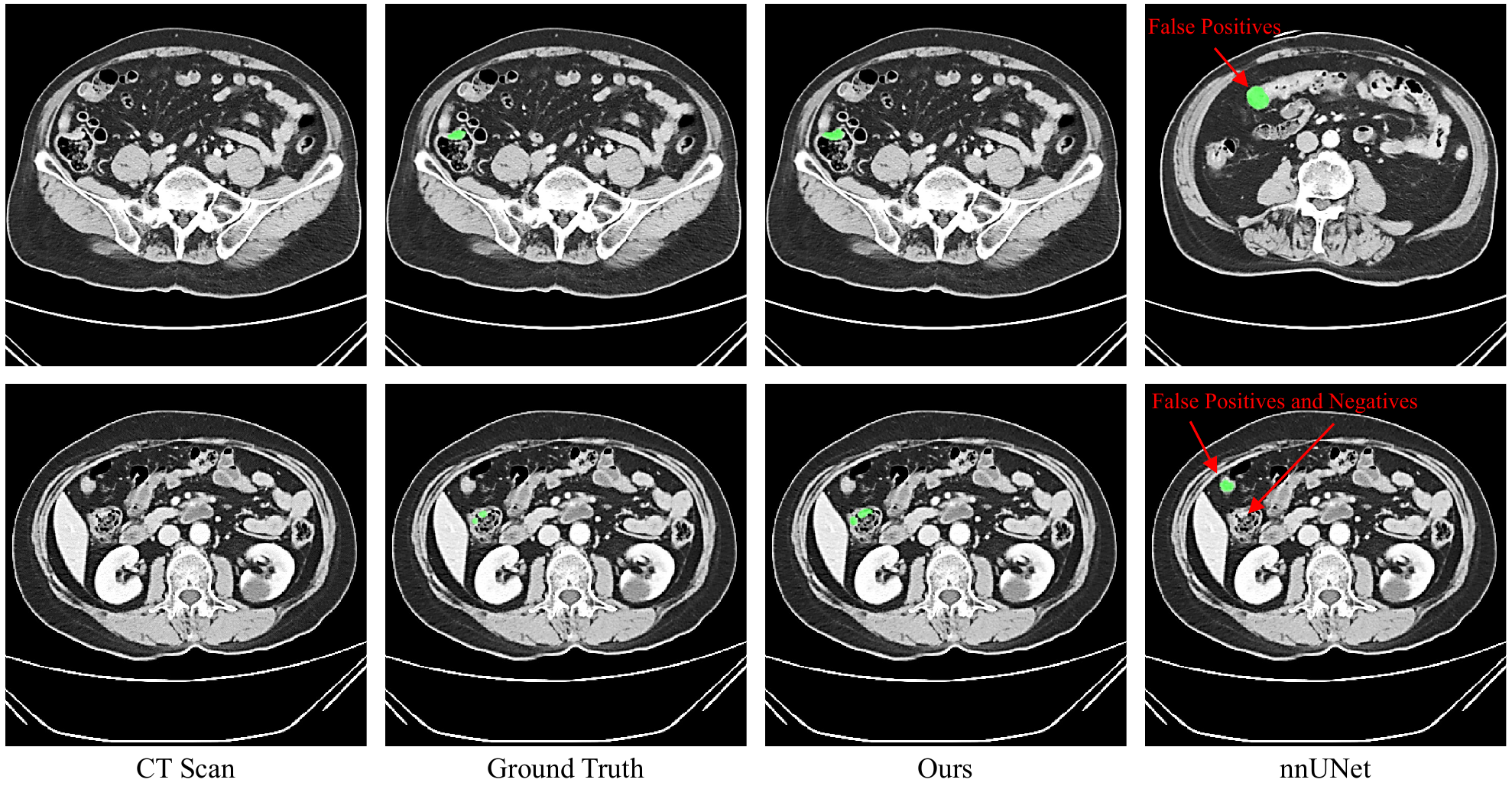}
  \caption{Comparative Visualization: Zero-shot ZePT versus Fully-Supervised nnUNet~\cite{isensee2021nnu} on real-world colon tumor segmentation dataset. Illustrated are two cases where ZePT successfully detects and segments colon tumors, contrasting with the fully-supervised nnUNet~\cite{isensee2021nnu}, which fails in these instances.
  }
  \label{fig:ZerovsFS}
\end{figure*}

\begin{figure*}
  \centering
  \includegraphics[width=\linewidth]{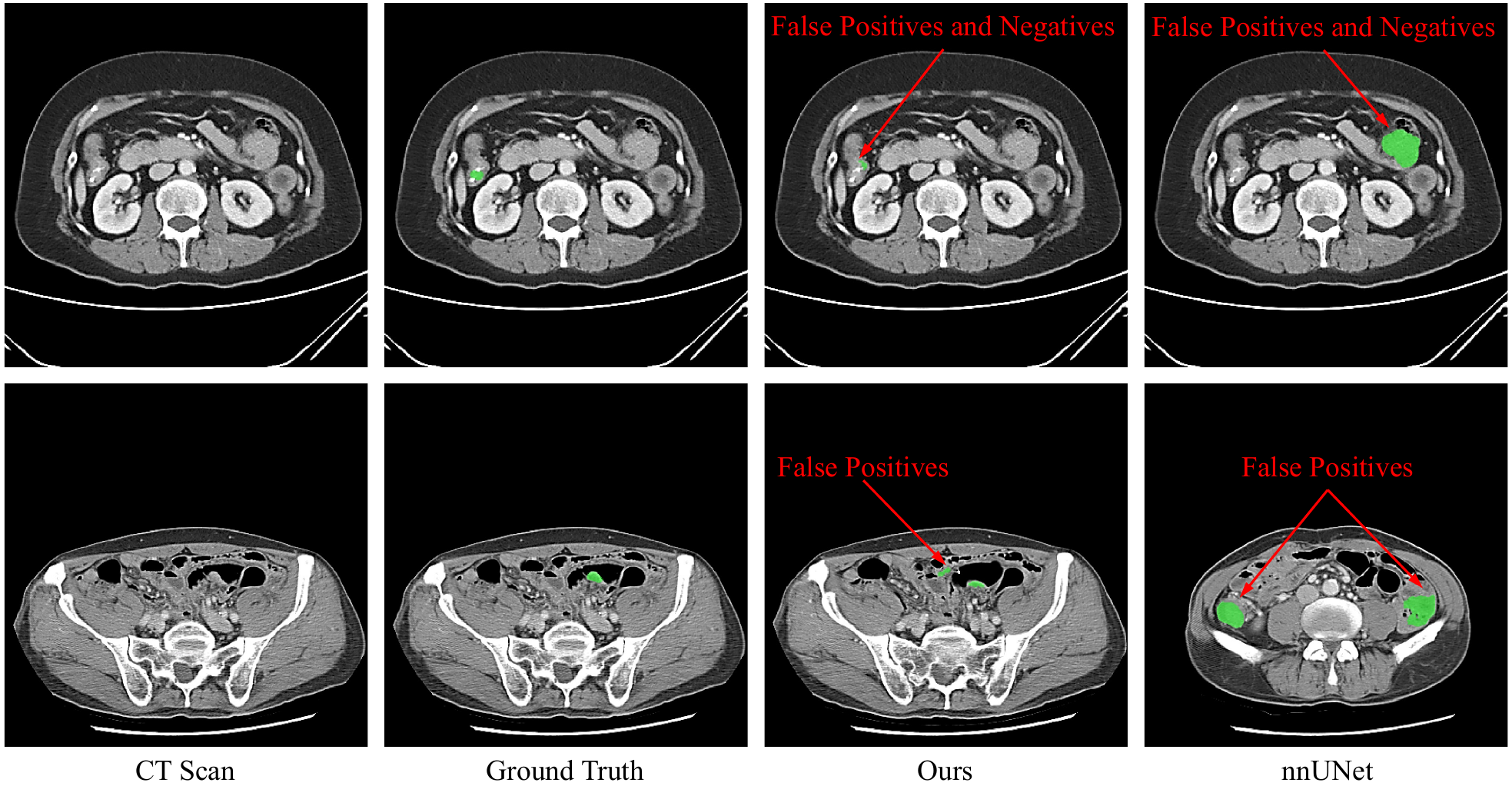}
  \caption{Failure Case Visualizations. Displayed are two examples where both zero-shot ZePT and fully-supervised nnUNet~\cite{isensee2021nnu} struggle due to the vague and indistinct characteristics of tumor areas.
  }
  \label{fig:failurecases}
\end{figure*}

And another important hyper-parameter is the number of queries. As mentioned in many previous studies~\cite{cheng2021per,cheng2022masked,yuan2023devil}, the number of queries should be larger than the possible/useful classes in the data, which depends heavily on the data and the task. In our task setup, for the segmentation of $25$ organs, we assign a fundamental query to each organ. Taking into account the actual occurrences of diseases associated with these organs, we use $20$ advanced queries for tumor segmentation. Consequently, in theory, our ZePT model is capable of identifying up to $20$ distinct tumor or lesion types. This configuration is flexible and can be modified according to the unique demands of various tasks.

\begin{table*}
  \Huge
  \centering
  \resizebox{\linewidth}{!}
  {
  \begin{tabular}{@{}l|cc|c|cc|cc|c|c@{}}
    \toprule
    \multicolumn{1}{l}{\multirow{2}{*}{Method}} & \multicolumn{2}{|c|}{Task03 Liver} & Task06 Lung Tumor & \multicolumn{2}{c|}{Task07 Pancreas} & \multicolumn{2}{c|}{Task08 Hepatic Vessel} & Task09 Spleen & Task10 Colon Tumor\\
    \cline{2-10}     
     & Organ DSC$\uparrow$ & Tumor DSC$\uparrow$ & DSC$\uparrow$ & Organ DSC$\uparrow$ & Tumor DSC$\uparrow$ & Organ DSC$\uparrow$ & Tumor DSC$\uparrow$ & DSC$\uparrow$ & DSC$\uparrow$ \\
    \hline
     nnUNet~\cite{isensee2021nnu}      & 94.57 & 58.22      & 66.57         & 80.06 & 50.45        & 63.29 & 68.20           & 96.53       & 50.07 \\
     Swin UNETR~\cite{tang2022self}    & 94.14 & 57.93      & 68.90         & 80.18 & 52.54        & 62.37 & 68.63           & 95.86       & 50.55 \\
     Universal~\cite{liu2023clip}      & 96.53 & 71.92      & 67.11         & 82.75 & 60.83        & 62.64 & 69.47           & 96.75       & 62.15 \\
    \hline
    ZePT (fully-supervised)    & \textbf{97.22} & \textbf{72.95}      & \textbf{69.07} 
    & \textbf{86.23} & \textbf{62.10}       & \textbf{64.39}  & \textbf{70.65}                 
    & \textbf{97.04}                & \textbf{64.87}   \\
    \bottomrule
  \end{tabular}
  }
  \caption{Benchmark on MSD validation dataset. We compare the fully-supervised ZePT with leading baselines, including nnUNet~\cite{isensee2021nnu}, Swin UNETR~\cite{tang2022self}, and Universal~\cite{liu2023clip} (previously ranked first on the MSD leaderboard), using 5-fold cross-validation on the MSD dataset. The fully-supervised ZePT demonstrated superior segmentation performance overall, particularly in segmenting the pancreas ($+3.48\%$), pancreatic tumors ($+1.27\%$), and colon tumors ($+2.72\%$).
  }
  \label{tab:fullySZePT}
\end{table*}

\subsection{Comparisons between Different Settings.}
\noindent \textbf{Zero-Shot vs. Fully-Supervised.}
While ZePT demonstrates outstanding performance in zero-shot tumor segmentation, its DSC scores are still lower compared to fully supervised models trained with labels of those unseen tumors. For instance, we trained the robust fully supervised nnUNet~\cite{isensee2021nnu} model on the collected real-world colon tumor segmentation dataset, achieving a average DSC of $58.30\%$. This exceeds ZePT's zero-shot colon tumor segmentation performance, which stands at a DSC of $36.23\%$, by a margin of $22.07$ percentage points. Despite this comparison being somewhat unfair, it highlights the substantial room for improvement in zero-shot learning for extremely challenging tasks like tumor segmentation. Notably, ZePT does not fall short in all aspects against the fully supervised nnUNet~\cite{isensee2021nnu}. Our experiments indicate that the fully supervised nnUNet~\cite{isensee2021nnu} tends to overfit the training data, resulting in missed detections and false positives in certain cases, particularly with rare tumor types. In contrast, ZePT consistently and accurately identifies and segments colon tumors in these cases. 
\cref{fig:ZerovsFS} illustrates two cases in which ZePT successfully segments colon tumors, whereas the fully supervised nnUNet~\cite{isensee2021nnu} produces false positives and negatives. This comparison highlights the substantial and promising potential of zero-shot learning to address the long-tail distribution challenge in medical imaging.

\cref{fig:failurecases} displays several failure cases from the real-world colon tumor segmentation dataset, characterized by exceedingly vague and indistinct tumor areas. In these instances, ZePT was unable to detect the tumors. It is important to note, however, that the fully supervised nnUNet~\cite{isensee2021nnu} model also struggled with these particularly challenging cases.


\noindent \textbf{Fully-Supervised ZePT.}
Additionally, we trained a fully-supervised version of ZePT to further evaluate its segmentation performance on seen organs and tumors. We follow the settings in~\cite{liu2023clip} and train a strong ZePT model on multiple public datasets. We then conducted a comparative analysis of the fully-supervised ZePT against established baselines, including nnUNet~\cite{isensee2021nnu}, Swin UNETR~\cite{tang2022self}, and Universal~\cite{liu2023clip}. Detailed comparisons based on 5-fold cross-validation on the MSD dataset are presented in~\cref{tab:fullySZePT}. The fully-supervised ZePT achieves overall better segmentation performance and offers substantial improvement in the tasks of segmenting pancreas ($+3.48\%$), pancreatic tumors ($+1.27\%$), and colon tumors ($+2.72\%$). This further demonstrates the novelty and superiority of ZePT's network architecture and training strategy. Notably, ZePT enhances segmentation performance significantly for seen organs and tumors, outperforming previous fully-supervised methods. Furthermore, it exhibits a remarkable ability for zero-shot tumor segmentation, a feature absent in traditional fully-supervised models. These strengths emphasize the importance and potential of ZePT.

\subsection{Differences Between ZePT and Existing Zero-Shot Medical Image Segmentation Methods}
Research on zero-shot segmentation models is scarcely explored within the medical imaging domain, primarily due to the intricacies involved in medical image segmentation tasks. Early attempts in this area include~\cite{ma2021zero,bian2021domain}.
Ma~\textit{et al.}~\cite{ma2021zero} proposed a zero-shot CNN model which utilizes two adjacent slices, instead of the target slice, as the input data of deep neural network to predict the brain tumor area in the target slice. This method has the potential to reduce the annotation workload, allowing doctors to only label a subset of the slices. Nevertheless, it requires tumor annotations for training and is capable of segmenting targets only when adjacent slices and their labels are supplied during the training process. Bian~\textit{et al.}~\cite{bian2021domain} introduced an annotation-efficient approach based on zero-shot learning for medical image segmentation. This method leverages the information in data of an existing image modality with detailed annotations and transfer the learned semantics to the target segmentation task with a new image modality. Therefore, their approach more closely resembles Domain Adaptation. These existing methods have yet been definitively proven to have the capability to segment multiple tumors in a strictly "zero-shot" manner. To the best of our knowledge, ZePT represents the first method capable of achieving zero-shot pan-tumor segmentation.

\subsection{Discussions on Future Works.}
\noindent \textbf{Improving Zero-Shot Tumor Segmentation Performance.}
The analysis of data from~\cref{tab:MSD} and~\cref{tab:abondataset} indicates that the model excels in zero-shot segmentation for unseen tumor categories that exhibit visual features similar to those of seen tumor categories. This suggests that simulating lesion features akin to unseen tumor categories during training, and directing the model to emphasize these features, could markedly improve its zero-shot segmentation capabilities. Consequently, future studies could investigate the use of diffusion-based models for simulating diverse tumor lesions' visual features and incorporating them into the training regime, which would potentially augment the model's effectiveness in zero-shot tumor segmentation.

\noindent \textbf{Adaptation to Diverse Imaging Modalities.} 
This paper primarily explores the zero-shot tumor segmentation challenge, without delving into addressing the differences between various imaging modalities. Consequently, in line with prior research~\cite{liu2023clip,chen2023towards} on creating universal segmentation models for various organs and tumors, our experiments and analyses were solely conducted using CT images. However, our approach is, in theory, adaptable and could potentially be applied to other 3D medical imaging modalities, including MRI and ultrasound. We aim to investigate this potential in future studies.

\noindent \textbf{Expanding Data Collection to Encompass a Broader Spectrum of Tumor Types for Evaluation.} 
In our research, we assembled a dataset of $388$ patients with colon tumors and utilized most of the publicly available tumor datasets to develop and evaluate the ZePT model's zero-shot performance. As previously noted, ZePT theoretically possesses the ability to segment a diverse array of unseen tumor types, beyond the scope of currently available datasets. To this end, we are actively compiling a more comprehensive dataset that includes a wider variety of tumor types, aiming to further assess ZePT's capabilities. Although the pervasive issue of data scarcity continues to challenge medical AI model development, ZePT represents a significant stride in overcoming this hurdle through zero-shot learning.







\end{document}